\documentclass[]{fairmeta}

\usepackage{amsmath,amsfonts,bm}

\def\eqref#1{equation~\ref{#1}}

\def\1{\bm{1}}

\def\rva{{\mathbf{a}}}
\def\rvb{{\mathbf{b}}}
\def\rvc{{\mathbf{c}}}

\def\rvh{{\mathbf{h}}}

\def\rvx{{\mathbf{x}}}

\def\rvz{{\mathbf{z}}}

\def\rmW{{\mathbf{W}}}

\DeclareMathAlphabet{\mathsfit}{\encodingdefault}{\sfdefault}{m}{sl}
\SetMathAlphabet{\mathsfit}{bold}{\encodingdefault}{\sfdefault}{bx}{n}

\usepackage{makecell}
\usepackage{enumitem}
\usepackage{colortbl}
\usepackage{xspace}
\usepackage{wrapfig}
\usepackage{amsmath}
\usepackage{amssymb}
\usepackage{mathtools}
\usepackage{amsthm}

\definecolor{RowHighlight}{gray}{0.9}
\newcommand{\llname}{LLM Pretraining with Continuous Concepts}
\newcommand{\lname}{Continuous Concept Mixing\xspace}
\newcommand{\mnamefirst}{Continuous Concept Mixing (CoCoMix)\xspace}
\newcommand{\mname}{Continuous Concept Mixing (CoCoMix)\xspace}
\newcommand{\mmname}{Continuous Concept Mixing\xspace}
\newcommand{\sname}{CoCoMix\xspace}

\title{\llname}

\author[1,2,\dagger]{Jihoon Tack}
\author[1]{Jack Lanchantin}
\author[1]{Jane Yu}
\author[1]{Andrew Cohen}
\author[1]{Ilia Kulikov}
\author[1]{Janice Lan}
\author[1,3,\dagger]{Shibo Hao}
\author[1]{Yuandong Tian}
\author[1]{Jason Weston}
\author[1]{Xian Li}

\affiliation[1]{FAIR at Meta}
\affiliation[2]{KAIST}
\affiliation[3]{UC San Diego}

\contribution[\dagger]{Work done during an internship at Meta}

\abstract{

Next token prediction has been the standard training objective used in large language model pretraining. Representations are learned as a result of optimizing for token-level perplexity. We propose \mname, a novel pretraining framework that combines discrete next token prediction with continuous concepts. Specifically, \sname predicts ``continuous concepts'' learned from a pretrained sparse autoencoder and mixes them into the model's hidden state by interleaving with token hidden representations. Through experiments on multiple benchmarks, including language modeling and downstream reasoning tasks, we show that \sname is more sample efficient and consistently outperforms standard next token prediction, knowledge distillation and inserting pause tokens. We find that combining both concept learning and interleaving in an end-to-end framework is critical to performance gains. Furthermore, \sname enhances interpretability and steerability by allowing direct inspection and modification of the predicted concept, offering a transparent way to guide the model’s internal reasoning process.

}

\date{\today}
\correspondence{Jihoon Tack at \email{jihoontack@kaist.ac.kr}, Xian Li at \email{xianl@meta.com}}

\metadata[Code]{\url{https://github.com/facebookresearch/RAM/tree/main/projects/cocomix}}

\begin{document}

\maketitle

\section{Introduction}
\label{sec:intro}

Recent progress in large language models (LLMs) has revolutionized natural language processing \citep{brown2020language,dubey2024llama} and thus became a core technology in various real-world applications, such as coding assistants \citep{roziere2023code}, search engines \citep{xuan2023evaluation}, and personal AI assistants \citep{gao2023assistgpt}. Central to these breakthroughs is the simple paradigm of next token prediction, which leverages massive amounts of unlabeled text to uncover rich linguistic patterns \citep{radford2018improving,radford2019language}. However, natural language tokens are often superficial (e.g., function words like ``the'' or ``a''), necessitating substantial training for models to acquire high-level reasoning and conceptual understanding while also hindering their ability to tackle long-horizon tasks such as planning \citep{lecun2022path,bachmann2024pitfalls}.

To tackle this issue, recent studies have investigated methods that go beyond token-level signals by leveraging richer information to train models. For instance, some approaches target more expressive prediction objectives, such as predicting multiple tokens at once to better capture semantic relationships \citep{gloeckle2024better,liu2024deepseek}, while others augment the input with rich signals, e.g., self-generated thought tokens \citep{zelikman2024quiet}, or fixed pause tokens \citep{goyal2024think} prior to next token prediction. Moreover, emerging evidence suggests that LLMs inherently encode high-level concepts and reasoning processes in their latent representations \citep{deng2023implicit,yang2024large}, indicating replacing discrete language tokens with continuous latent representations has promise in improving reasoning efficiency \citep{hao2024training}. While token-level modeling remains important for coherent text generation, the key challenge is to enrich or supplement these natural language tokens so that LLMs can learn more abstract reasoning abilities and long-range dependencies.

This raises a key question: can we augment the next token prediction objective to explicitly model concepts in a latent representation space, thereby bridging semantic abstraction and fine-grained token-level guidance?

To this end, we draw inspiration from recent findings that \emph{Sparse Autoencoders (SAEs)} can effectively isolate meaningful latent features in LLMs by capturing the high-level semantic concepts \citep{cunningham2023sparse, bricken2023monosemanticity}. As SAEs are trained to reconstruct the model's hidden state with sparsity constraints, it encourages focusing on a compact set of concept dimensions \citep{templeton2024scaling}. This makes it possible to highlight the pretrained model’s concepts---the core semantic directions that underpin model prediction---while avoiding unnecessary features. 

\begin{figure*}[t!]
\centering
\includegraphics[width=\textwidth]{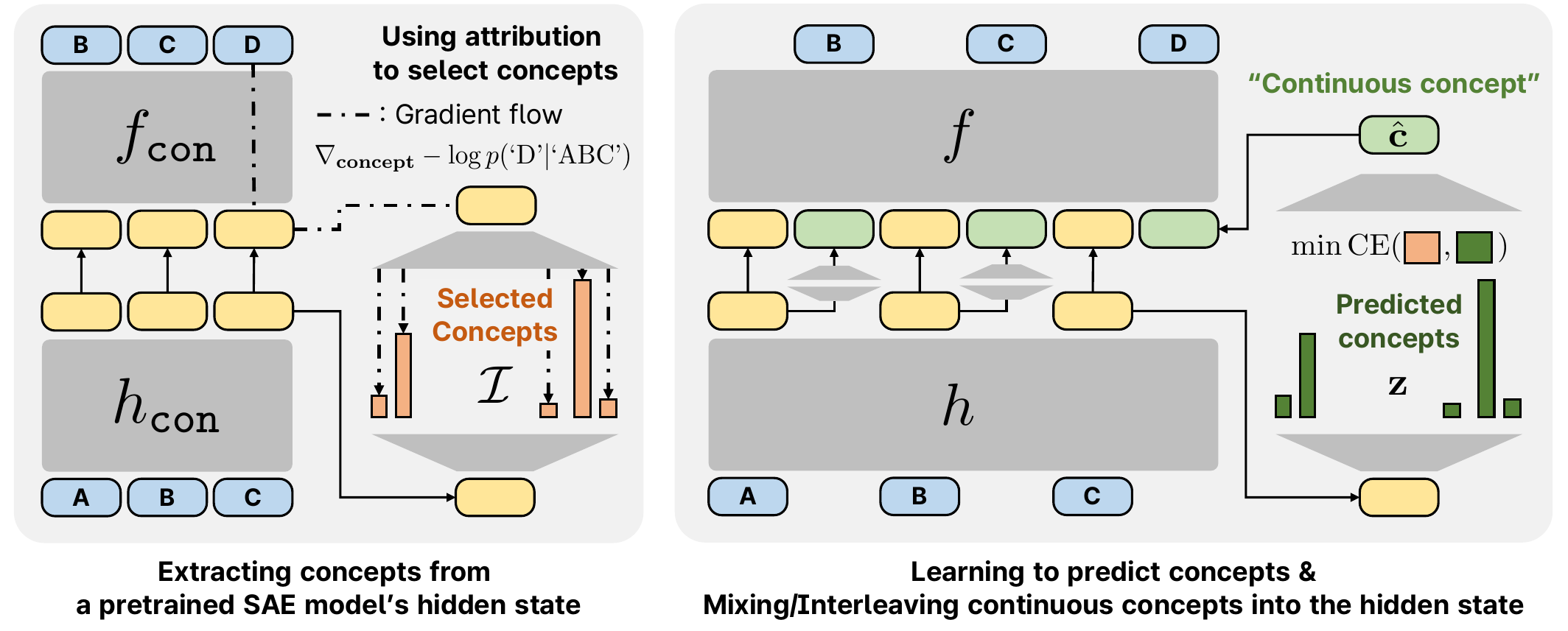}
\caption{\textbf{Overview of \sname.} We use an SAE to extract concepts from a pretrained model's hidden state $h_\mathtt{con}$ and then select important concepts based on the attribution score (i.e., measuring the influence on the output). These selected concepts are used as labels $\mathcal{I}$ for concept prediction by minimizing the cross-entropy loss $\mathrm{CE}(\cdot,\cdot)$.
The predicted concepts $\rvz$ are then compressed into a compact vector, forming a continuous concept $\rvc$, which is mixed into the model’s hidden state by interleaving with token hidden representations. We demonstrate that \sname is more sample efficient and outperforms standard next-token prediction and knowledge distillation baselines.
}
\label{fig:concept}
\end{figure*}

We propose \mnamefirst, a novel and effective language modeling framework using continuous concepts. Specifically, we extract semantic concepts using a pretrained SAE and select the most influential ones based on attribution scores, which quantify each concept’s influence on the model’s output. The model is then trained to predict these selected concepts from its hidden state using a cross-entropy loss. Once multiple concepts are predicted, we compress them into a single \emph{continuous concept} and \emph{mix} (or insert) into the hidden states by interleaving with token embeddings, thereby directly contributing to the next token prediction. An additional benefit is that one can probe or analyze the predicted concepts, enabling controllable generation and improving model interpretability. An overview of \sname is shown in \autoref{fig:concept}.

We demonstrate the efficacy of \sname through extensive evaluations across multiple language modeling benchmarks and pretraining model sizes, ranging from million-scale to billion-scale parameter sizes. For instance, when applied to a 1.38B-sized model, \sname achieves comparable performance with the next token prediction with 21.5\% fewer training tokens. Moreover, \sname demonstrates substantial improvements in weak-to-strong supervision scenarios, where concepts extracted from a small model can even be used as ground truth labels to supervise the training of a larger model. Finally, we show that the insertion of compressed concept vectors enables us to steer and control the model by probing the predicted concept during generation.

\section{\sname: \lname}
\label{sec:method}

In this section, we present \mnamefirst, a framework that extends next-token prediction with continuous concepts. We first review our problem setup of interest and the concept extraction process using a sparse autoencoder (SAE) in Section~\ref{sec:prelim}. Then, we describe the concept selection framework and explain how the model predicts and interleaves these concepts into its hidden state to improve language modeling in Section~\ref{sec:method_cocomix}.

\subsection{Prelim: Problem Setup and Sparse Autoencoder}
\label{sec:prelim}

\textbf{Problem setup.} 
We consider a language-modeling task over a vocabulary $\mathcal{V}$, with a training corpus $\mathcal{D} \subseteq \mathcal{V}^*$, where each sequence $\rvx=(x_{1},x_{2},\ldots)\in \mathcal{D}$ is drawn i.i.d.\ from the data distribution. We require a pretrained language model for concept extraction  $\mathcal{M}_\mathtt{con}=f_\mathtt{con}\circ h_\mathtt{con}$ which predicts next token probability $\mathcal{M}_\mathtt{con}(x_{t} \mid x_{<t})$. Here, we extract the continuous concepts $\rvc$ from the hidden state $h_\mathtt{con}$, where $h_\mathtt{con}$ outputs the hidden state at a chosen layer of depth $L_\mathtt{con}$, and $f_\mathtt{con}$ maps that representation to the next token distribution. We aim to train $\mathcal{M}$ to model the same autoregressive factorization $\mathcal{M}(\rvx) \;=\; \prod_{t=1}^{|\rvx|} \mathcal{M}(x_{t} \mid x_{<t}, \rvc_{<t})$, but by augmenting with continuous concepts $\rvc$.

\textbf{Sparse Autoencoder.}
We first propose to extract high-level concepts from a pretrained model's hidden state. We use an SAE, which maps the input to a high-dimensional sparse activation, with the objective of reconstructing the original input \citep{lee2006efficient}. When applied to an LLM, the SAE decomposes the hidden state into multiple dimensions, each of which can be viewed as a distinct concept capturing semantically meaningful features \citep{yun2021transformer,bricken2023monosemanticity}. We consider the TopK SAE \citep{makhzani2014k}, which uses a TopK activation to enforce sparsity.

Formally, for a given input sequence $\rvx$ and corresponding pretrained model's hidden state at position $t$, denoted as $\rvh_t^{\mathtt{con}}=h_\mathtt{con}(\rvx)_t \in \mathbb{R}^{d_\mathtt{con}}$, SAE consists of a linear encoder $E: \mathbb{R}^{d_{\mathtt{con}}} \to \mathbb{R}^{C}$ and a linear decoder $D:\mathbb{R}^{C} \to \mathbb{R}^{d_\mathtt{con}}$, where $C$ is the dimension of the concept space. The reconstruction process of SAE is as:
\begin{equation*}
    \rvc_t^{\mathtt{pre}} = E\bigl(\rvh_t^\mathtt{con}\bigr), 
    ~~
    \rvc_t = \mathrm{TopK}\bigl(\rvc_t^{\mathtt{pre}}\bigr),
    ~~
    \widehat{\rvh}_t^\mathtt{con} = D\bigl(\rvc_t\bigr),
\end{equation*}
where $\rvc_t^{\mathtt{pre}}\in\mathbb{R}^{C}$ is the pre-activation concept vector, $\mathrm{TopK}(\cdot)$ zeros out all but the largest \(K_{\mathtt{SAE}}\) entries, and $\widehat{\rvh}_t^\mathtt{con}$ is the reconstruction. The SAE is trained by minimizing the reconstruction loss: $\lVert \rvh_t^\mathtt{con} - \widehat{\rvh}_t^\mathtt{con}\rVert_{2}^2$. By enforcing TopK sparsity, the SAE isolates the most critical dimensions in $\rvc$ that explain the pretrained model’s features. Here, each activated element of $\rvc_t$ is interpreted as a concept.

\subsection{\mmname}
\label{sec:method_cocomix}

Now, we describe the core training pipeline of \sname, which consists of a concept selection framework (refer to the left figure in \autoref{fig:concept}) and two training steps to learn and utilize continuous concepts (refer to the right figure in \autoref{fig:concept}). First, we select important concepts using the attribution score, which measures the influence of each concept on the output. Then, we propose predicting the selected concepts from the model's hidden state using cross-entropy loss, allowing the model to implicitly learn which concepts should be encoded as a hidden representation. Finally, we utilize the predicted concepts to create a ``continuous concept'', which is interleaved in the hidden states, enabling the model to explicitly learn to use both the continuous concepts as well as the token hidden states. Intuitively, the model selectively learns which concepts are useful for the next token prediction and how it should \emph{mix} them with the token representations.

\textbf{Target concept selection using attribution.}
While the extracted concept $\rvc_t$ represents the core features of the current input context, it may not directly indicate which concepts matter most for predicting the ground truth next token $x_{t+1}$  \citep{templeton2024scaling}. To this end, we propose utilizing \emph{attribution}, a method to measure the causal influence of each concept on the output \citep{baehrens2010explain,sundararajan2017axiomatic}. Specifically, the attribution score measures the influence based on the local linear approximation of the effect of changing the concept value. In this paper, we use a simple attribution score that measures the influence by multiplying the loss gradient with a given input \citep{simonyan2014very,shrikumar2016not}.

Concretely, for a given input $\rvx$ and corresponding concept $\rvc_t$, we define the attribution score $\rva_t\in\mathbb{R}^{C}$ as:
\begin{equation}
\rva_t \;=\; 
\rvc_t^{\mathtt{pre}} \,\odot\, \nabla_{\rvc_t} -\log f_{\mathtt{con}}\big(x_{t+1}|D(\rvc_{t}),\rvh_{<t}\big),
\end{equation}
where $\odot$ denotes element-wise multiplication. Note that we multiplied the pre-activation $\rvc_t^{\mathtt{pre}}$ (instead of $\rvc_t$) to capture non-activated concepts made by the TopK sparse activation. Based on the computed attribution score, we select salient concepts and incorporate them into language modeling.

\textbf{Predicting the selected concepts.} 
For a given attribution score $\rva_t$, we first select the indices of the concept that have a high score and use these as discrete labels for concept prediction. Let $\mathcal{I}=\{i_1,\ldots,i_{K_{\text{attr}}}\}$ be the set of indices corresponding to the top $K_{\text{attr}}$ values of $\rva_t$, and $\rvh_t=h(\rvx)_{t} \in \mathbb{R}^{d}$ be the model’s hidden state of the given input $\rvx$ at the same token position. Then, we learn to predict these labels using a linear prediction head $M$ that outputs logit $\rvz_t=M(\rvh_t) \in \mathbb{R}^{C}$ by minimizing the following cross-entropy loss $\mathrm{CE}(\cdot, \cdot)$ is as follows:
\begin{equation}
    \mathcal{L}_{\mathtt{concept}}(\rva_t) 
    \;=\;
    \frac{1}{K_{\text{attr}}}\sum_{i \in \mathcal{I}} \mathrm{CE}~\!\bigl(\rvz_t,\; i\bigr).
    \label{eq:concept_pred}
\end{equation}

\textbf{Mixing continuous concepts with token embeddings.}
To encourage the model to internalize the concept more effectively, we propose ``mixing'' (i.e., interleaving) the predicted concept with the existing token hidden representations. As our model predicts multiple concepts at once, we propose to compress them into a compact vector through a learnable mapping, which we refer to as a continuous concept. This compressed vector is interleaved with token vectors.

Formally, for a given concept prediction logit $\rvz_t$, we sparsify the logit using $\mathrm{TopK}$ activation to predict the concepts, then compress them into a continuous concept vector $\hat{\rvc}_t \in \mathbb{R}^{d}$:
\begin{equation}
    \hat{\rvc}_t \;=\; \rmW\, \mathrm{TopK}(\rvz_t) + \rvb,
    \label{eq:compress}
\end{equation}
where $\rmW \in \mathbb{R}^{d \times C}$ and $\rvb \in \mathbb{R}^{d}$ project the TopK-sparse vector to a $d$-dimensional embedding. We then append $\hat{\rvc}_t$ to the model’s hidden sequence $\hat{\rvc}_t$ as $(\rvh_{1}, \hat{\rvc}_{1}, \ldots, \rvh_{t}, \hat{\rvc}_{t})$, which is fed into the remaining transformer blocks $f$. Note that this design not only improves the model’s performance but also offers interpretability and controllability through analyzing and steering the predicted concept $\rvz_t$, which can be probed or tuned during the model's generation process. Furthermore, by analyzing the weights of the compression layer $\rmW$, one can identify which concept is useful for the next token prediction. 

This approach shares similarities with intervention techniques in the mechanistic interpretability literature, which modify the hidden state by adding a concept vector to the original hidden state \citep{zou2023representation,wu2024reft}. However, unlike intervention methods that directly manipulate the hidden state, our approach treats the predicted concept as a separate unit of information by interleaving it in the hidden state. This design allows the model to process the concept independently, leveraging the pretrained model's internal reasoning reflected in the prediction. 

\textbf{Training objective.}
The final training objective for \sname combines the standard next token prediction loss and the concept prediction term as follows:
\begin{equation}
    \sum_{t=1}^{T-1} -\log f\bigl(x_{t+1}\mid \rvh_{\leq t}, \hat{\rvc}_{\leq t}\bigr)
    \;+\;
    \lambda \mathcal{L}_{\mathtt{concept}}(\rva_t),
    \label{eq:final}
\end{equation}
where $\lambda$ is a tunable coefficient.

\section{Experiments}
\label{sec:experiments}

\begin{figure*}[t]
\centering\small

\begin{subfigure}[b]{0.3275\textwidth}
    \centering
    \includegraphics[width=\textwidth,height=0.75\textwidth]{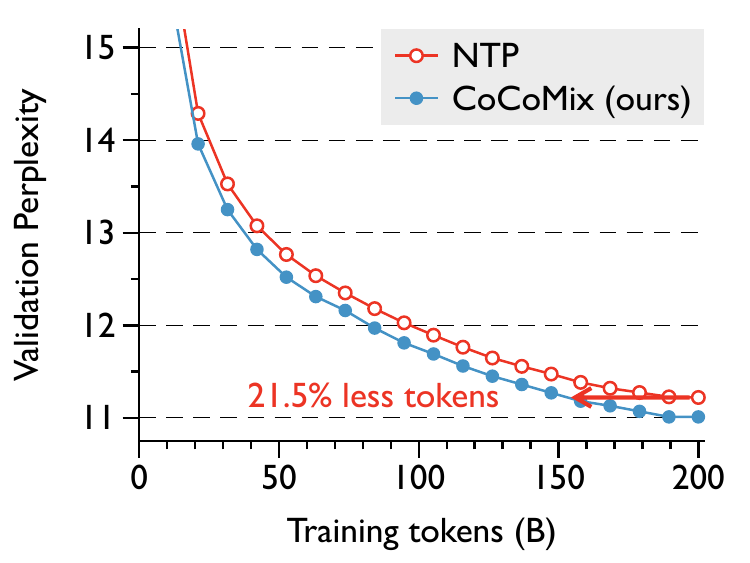}
    \caption{Validation perplexity}
\end{subfigure}
\hfill
\begin{subfigure}[b]{0.3275\textwidth}
    \centering
    \includegraphics[width=\textwidth,height=0.75\textwidth]{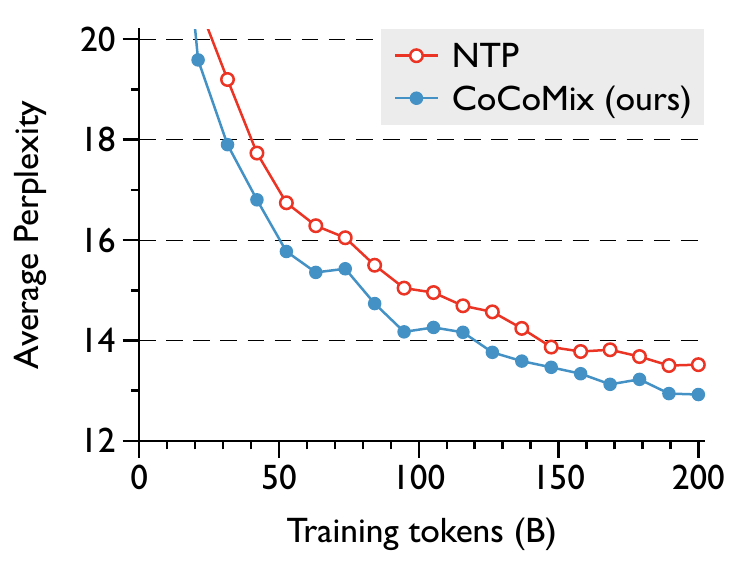}
    \caption{Average downstream task perplexity}
\end{subfigure}
\hfill
\begin{subfigure}[b]{0.3275\textwidth}
    \centering
    \includegraphics[width=\textwidth,height=0.75\textwidth]{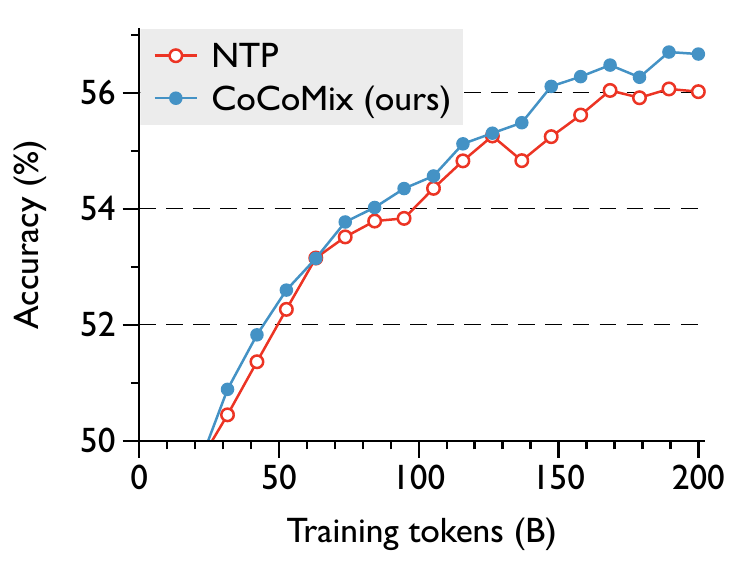}
    \caption{Average downstream task accuracy}
\end{subfigure}
\caption{
\textbf{\sname vs. NTP performance at different training checkpoints.} Each model contains a total of 1.38B parameters. Each model is trained on the OpenWebText dataset. For \sname, the concepts are extracted from a 124M-sized model (10$\times$ smaller than the base model). The plots show improvements in: (a) validation perplexity, (b) average perplexity on LAMBADA, WikiText-103, and (c) average accuracy on HellaSwag, PIQA, SIQA, Arc-Easy, and WinoGrande.
}
\label{fig:main_curve}
\end{figure*}

\begin{figure*}[t!]
\centering\small

\begin{subfigure}{0.245\textwidth}
    \includegraphics[width=\textwidth,height=0.65\textwidth]{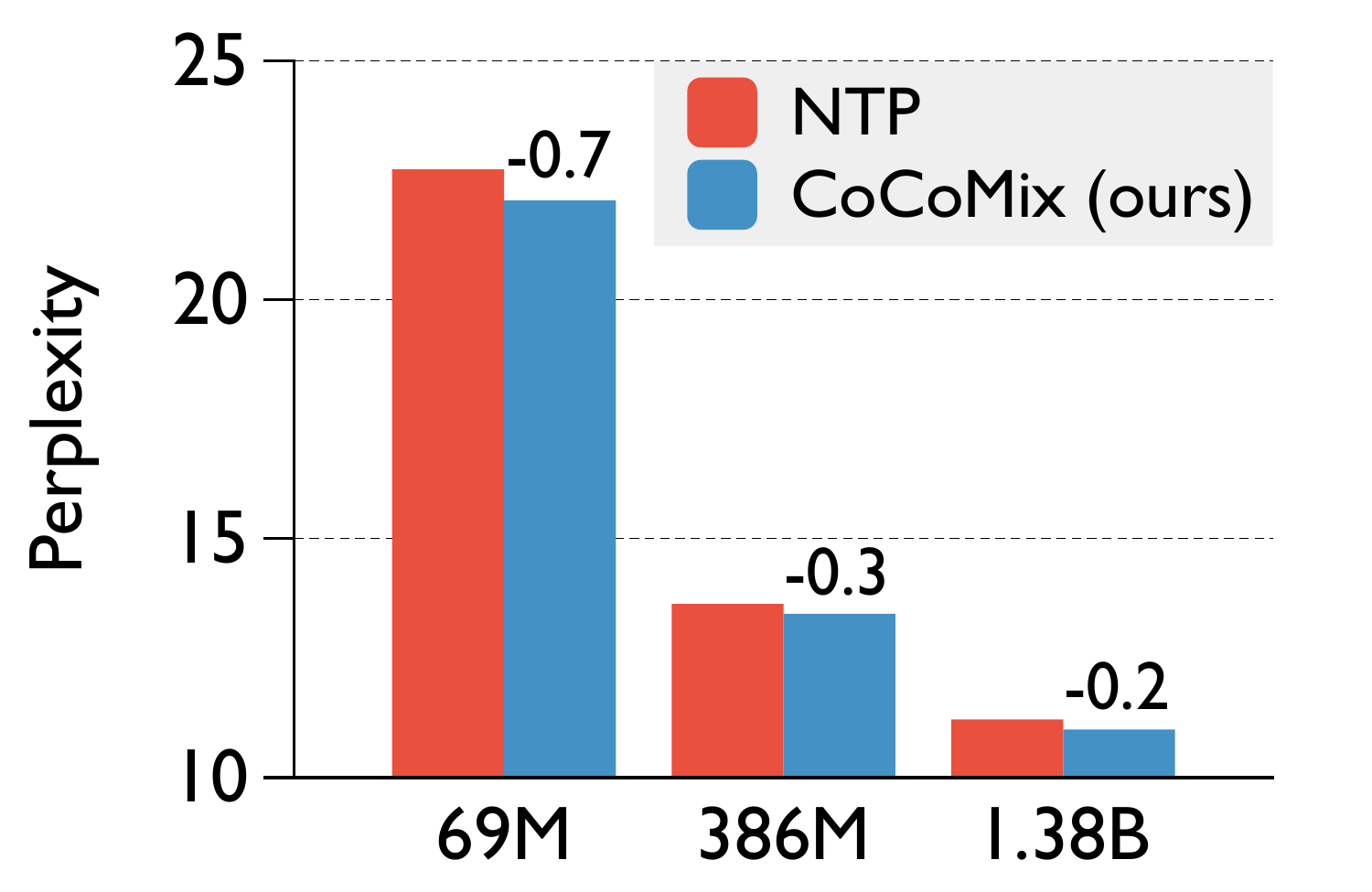}
    \caption{OpenWebText, PPL ($\downarrow$)}
\end{subfigure}
\begin{subfigure}{0.245\textwidth}
    \includegraphics[width=\textwidth,height=0.65\textwidth]{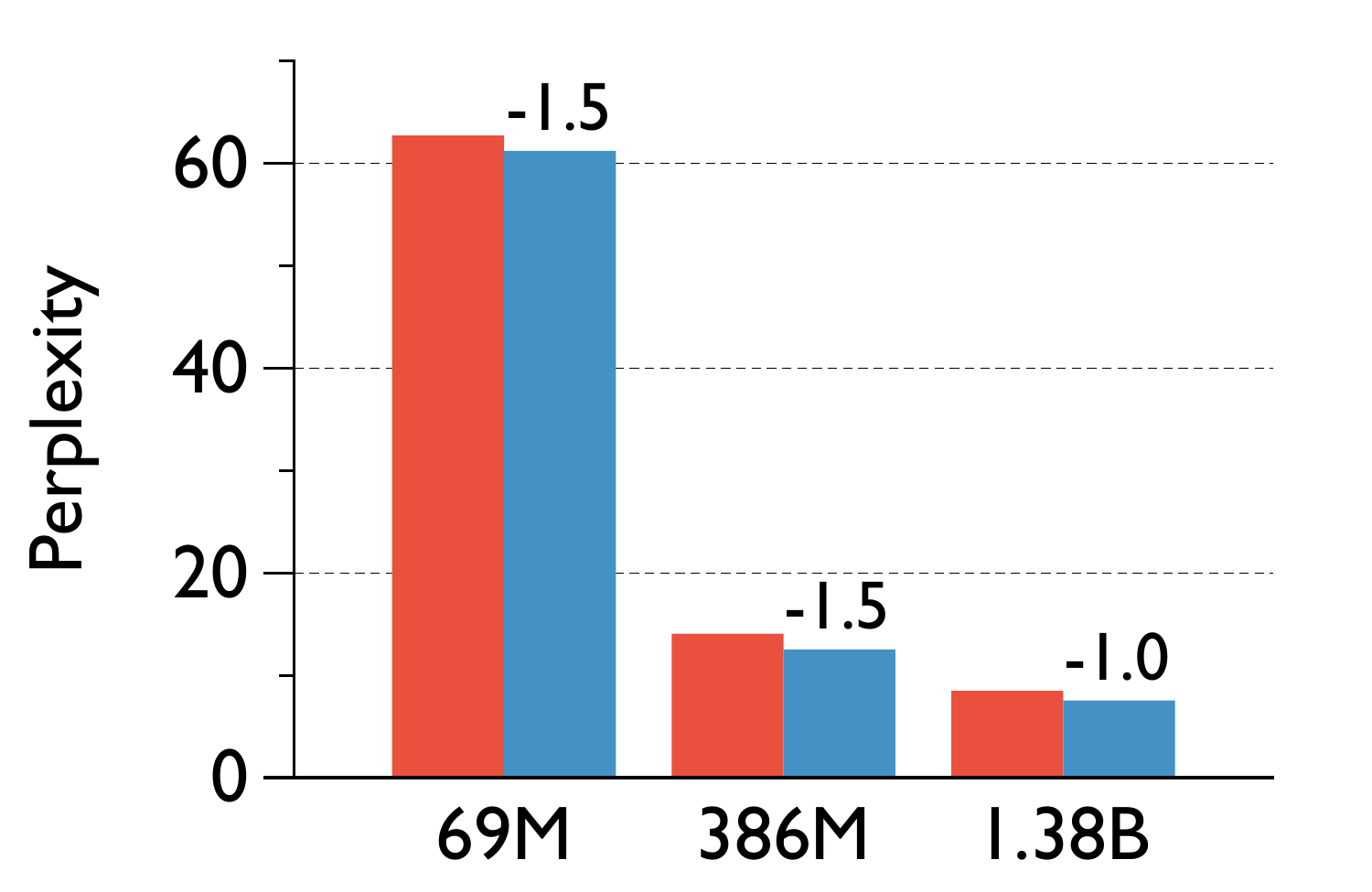}
    \caption{LAMBADA, PPL ($\downarrow$)}
\end{subfigure}
\begin{subfigure}{0.245\textwidth}
    \includegraphics[width=\textwidth,height=0.65\textwidth]{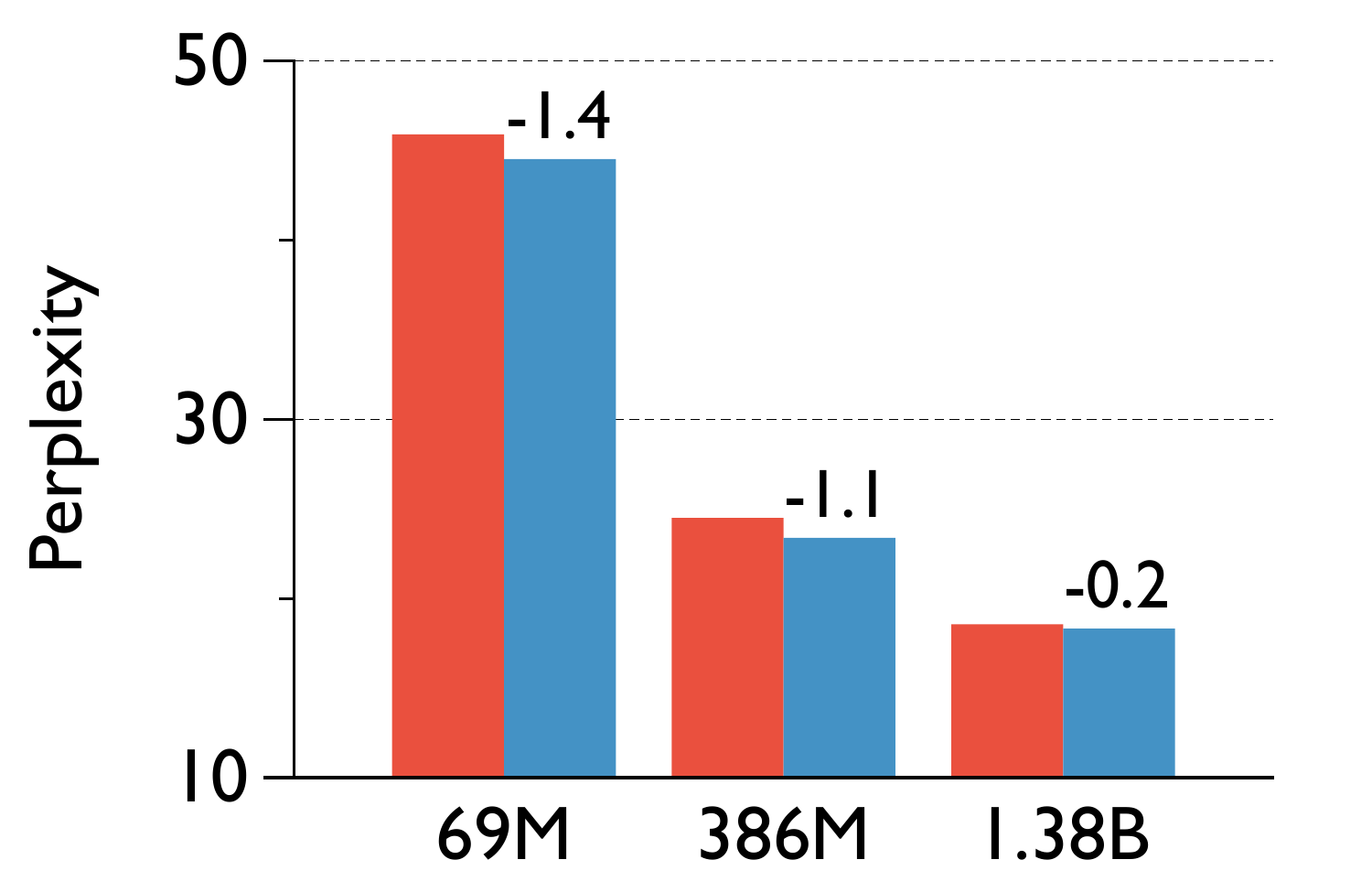}
    \caption{WikiText-103, PPL ($\downarrow$)}
\end{subfigure}
\begin{subfigure}{0.245\textwidth}
    \includegraphics[width=\textwidth,height=0.65\textwidth]{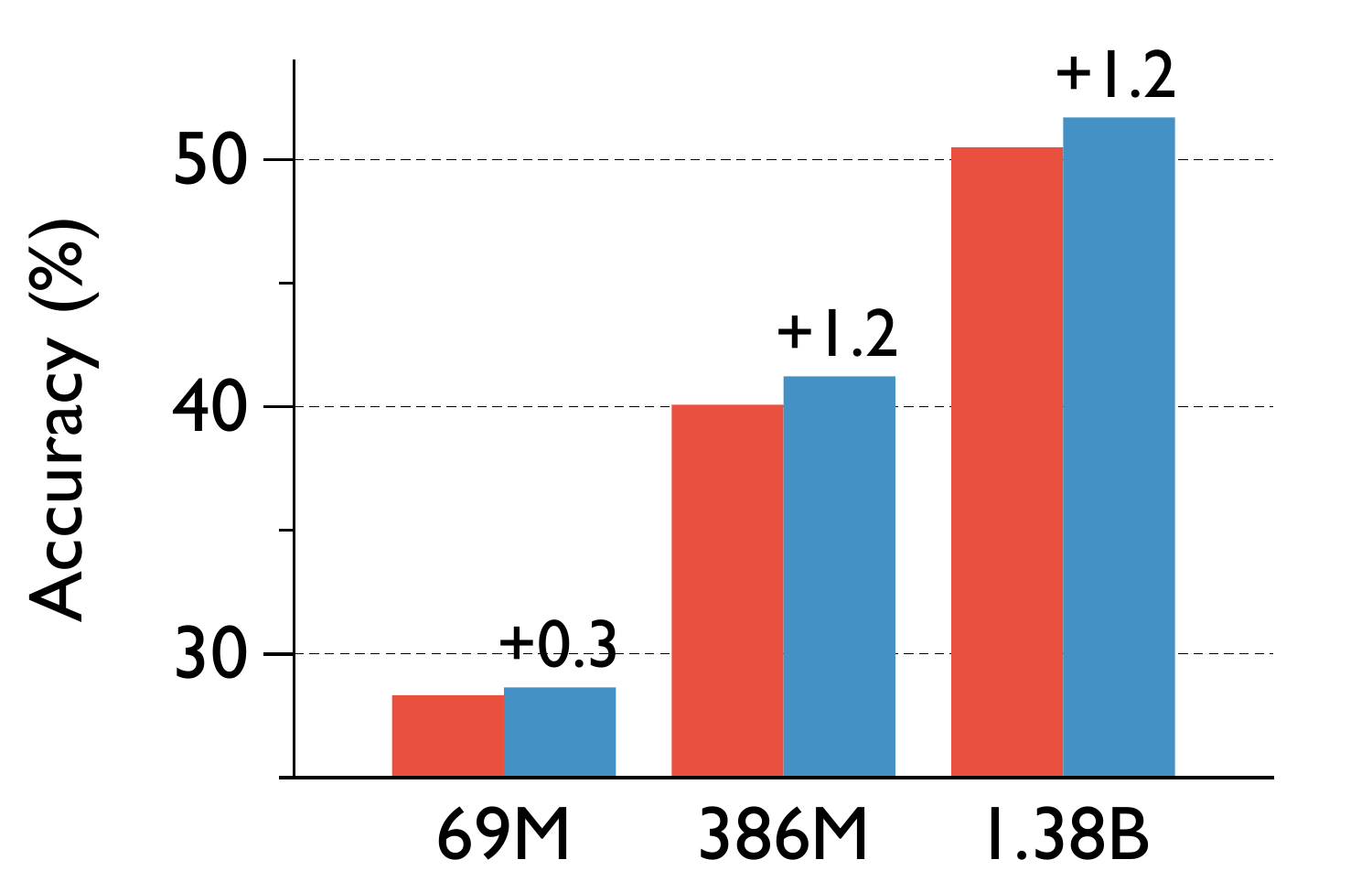}
    \caption{HellaSwag, Acc-n ($\uparrow$)}
\end{subfigure}
\begin{subfigure}{0.245\textwidth}
    \includegraphics[width=\textwidth,height=0.65\textwidth]{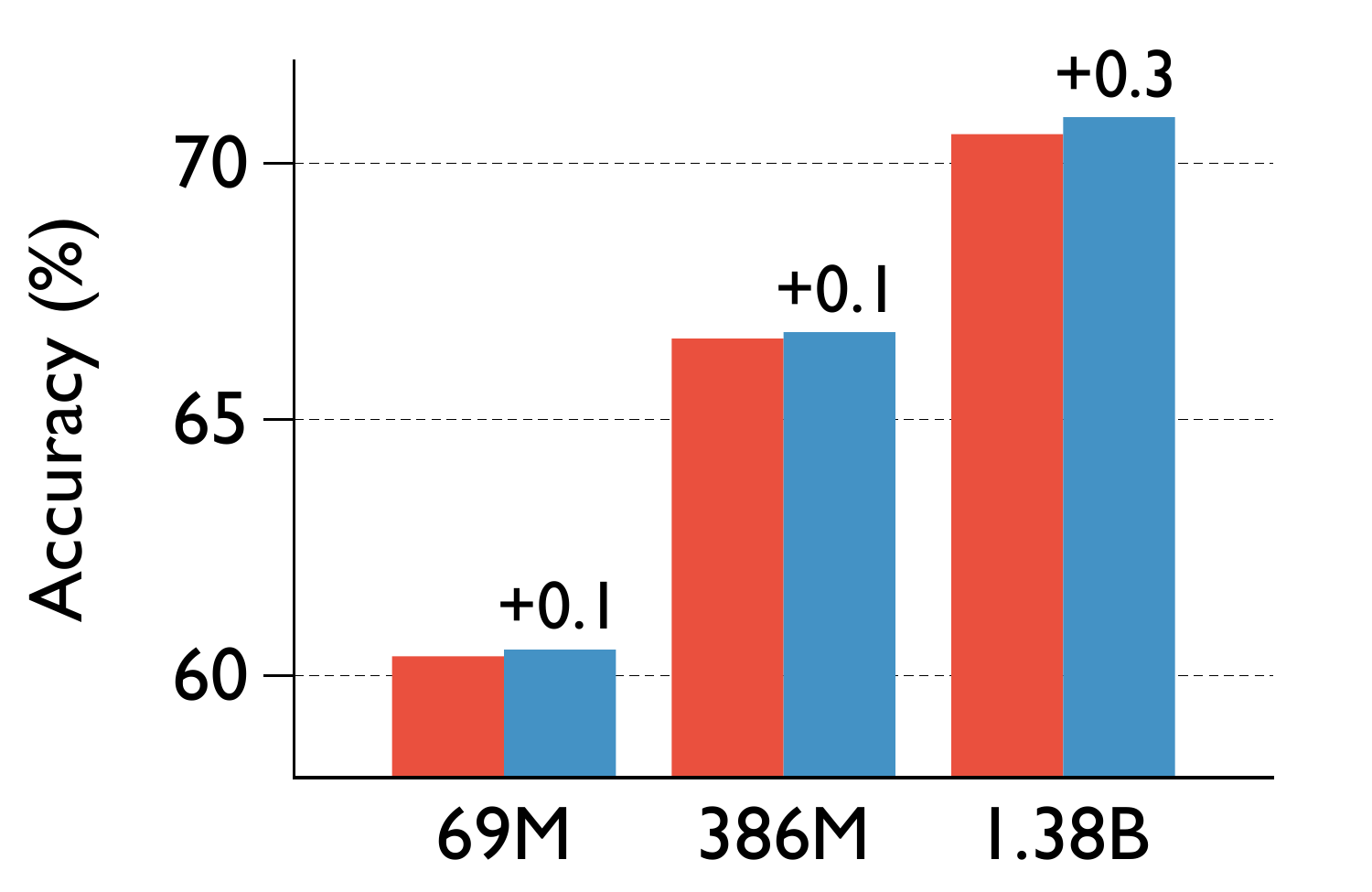}
    \caption{PIQA, Acc ($\uparrow$)}
\end{subfigure}
\begin{subfigure}{0.245\textwidth}
    \includegraphics[width=\textwidth,height=0.65\textwidth]{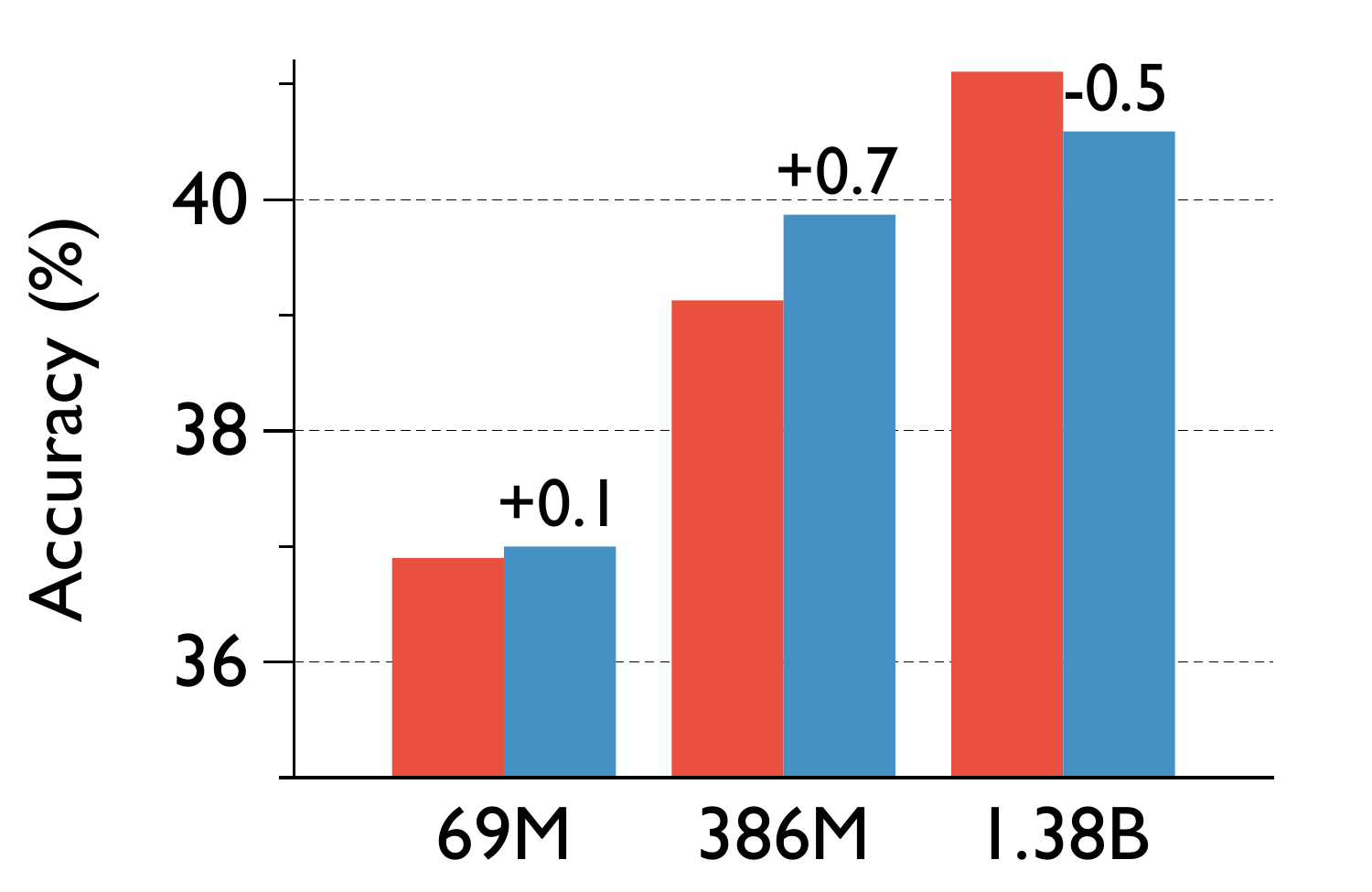}
    \caption{SIQA, Acc ($\uparrow$)}
\end{subfigure}
\begin{subfigure}{0.245\textwidth}
    \includegraphics[width=\textwidth,height=0.65\textwidth]{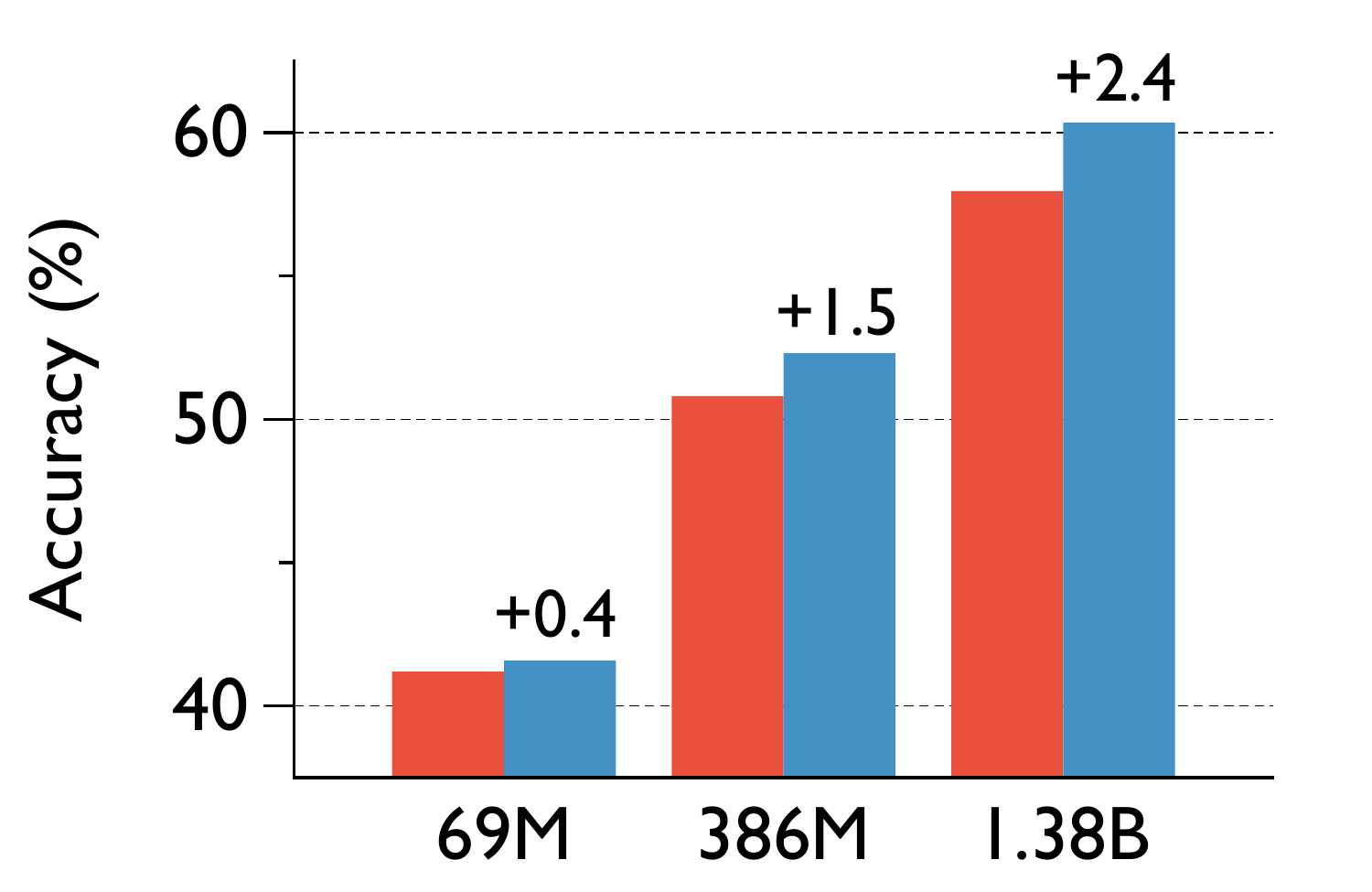}
    \caption{Arc-Easy, Acc ($\uparrow$)}
\end{subfigure}
\begin{subfigure}{0.245\textwidth}
    \includegraphics[width=\textwidth,height=0.65\textwidth]{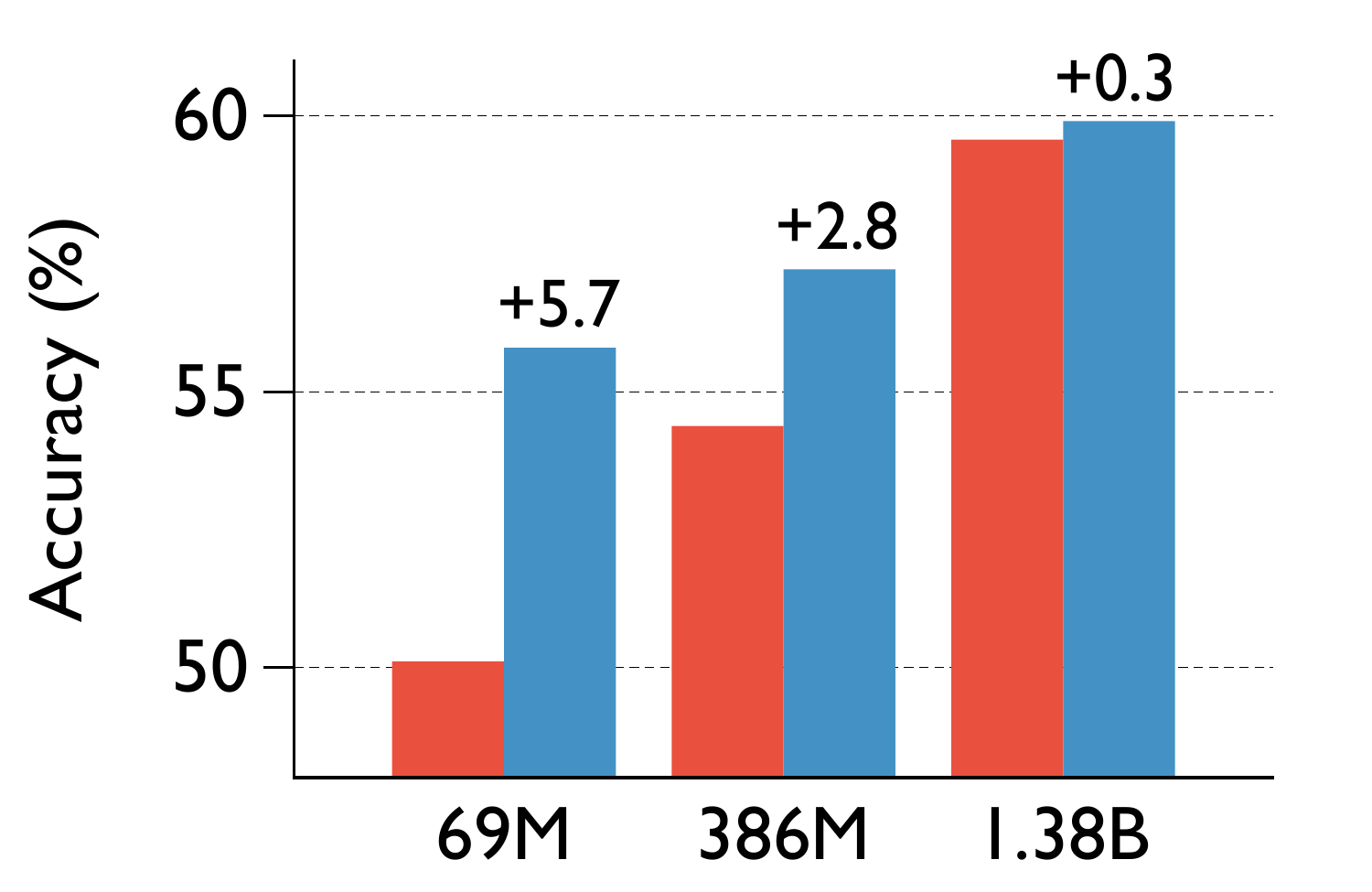}
    \caption{WinoGrande, Acc ($\uparrow$)}
\end{subfigure}
\caption{
\textbf{\sname vs. NTP performance at different model sizes.} We consider various model sizes, including 69M, 386M, and 1.38B parameters and train on 200B OpenWebText tokens. We evaluate the models on OpenWebText validation perplexity and downstream datasets LAMBADA, WikiText-103, HellaSwag, PIQA, SIQA, Arc-Easy, and WinoGrande.
}
\label{fig:downstream}
\end{figure*}

\begin{figure*}[t]
\centering\small
\begin{subfigure}{0.3275\textwidth}
    \includegraphics[width=\textwidth,height=0.75\textwidth]{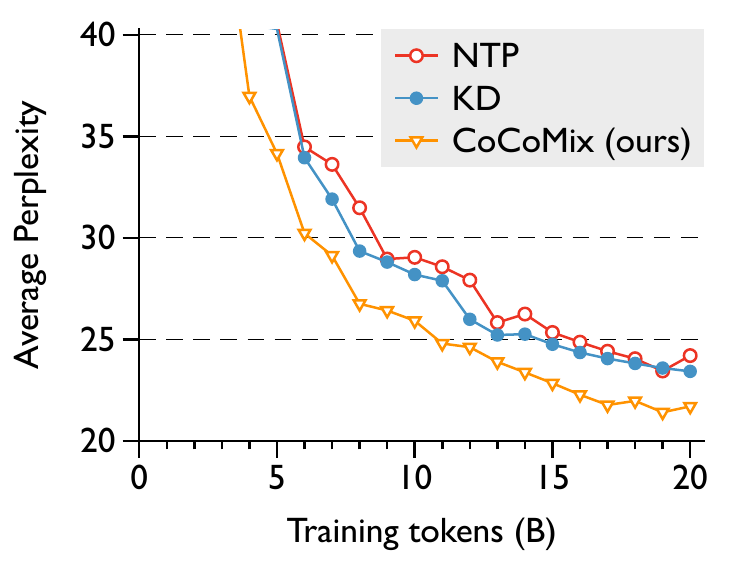}
    \caption{Weak-to-strong: Avg. Perplexity}
    \label{fig:weak_to_strong_perp}
\end{subfigure}
\hfill
\begin{subfigure}{0.3275\textwidth}
    \includegraphics[width=\textwidth,height=0.75\textwidth]{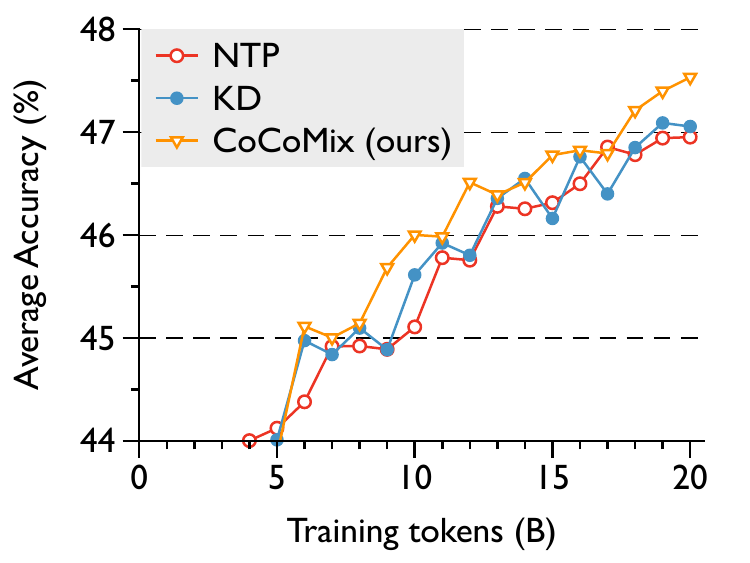}
    \caption{Weak-to-strong: Avg. Accuracy}
    \label{fig:weak_to_strong}
\end{subfigure}
\hfill
\begin{subfigure}{0.3275\textwidth}
    \includegraphics[width=\textwidth,height=0.75\textwidth]{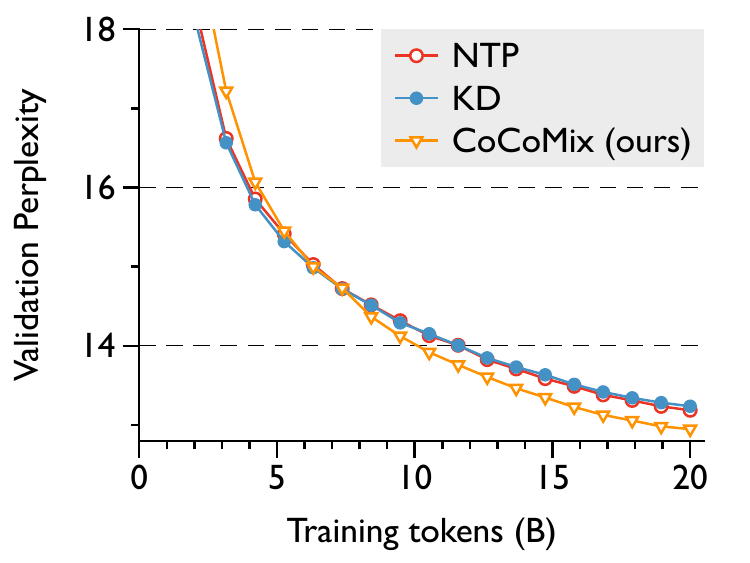}
    \caption{Distribution shift (OpenWebMath)}
    \label{fig:dist_shift}
\end{subfigure}
\caption{
\textbf{\sname vs. Knowledge Distillation (KD).} 
For our weak-to-strong supervision setup, we train a 386M model where the teacher of KD (or concept extraction for \sname) is a 124M-sized model: we report (a) average perplexity over OpenWebText, LAMABADA, and WikiText and (b) average accuracy over HellaSwag, PIQA, SIQA, Arc-Easy, and WiniGrande dataset. For (c) a distribution shift setup, we train all methods on OpenWebMath, a math specific pretraining corpus.
}
\label{fig:baseline}
\end{figure*}

\begin{figure*}[t]
    \centering
    \includegraphics[width=\linewidth]{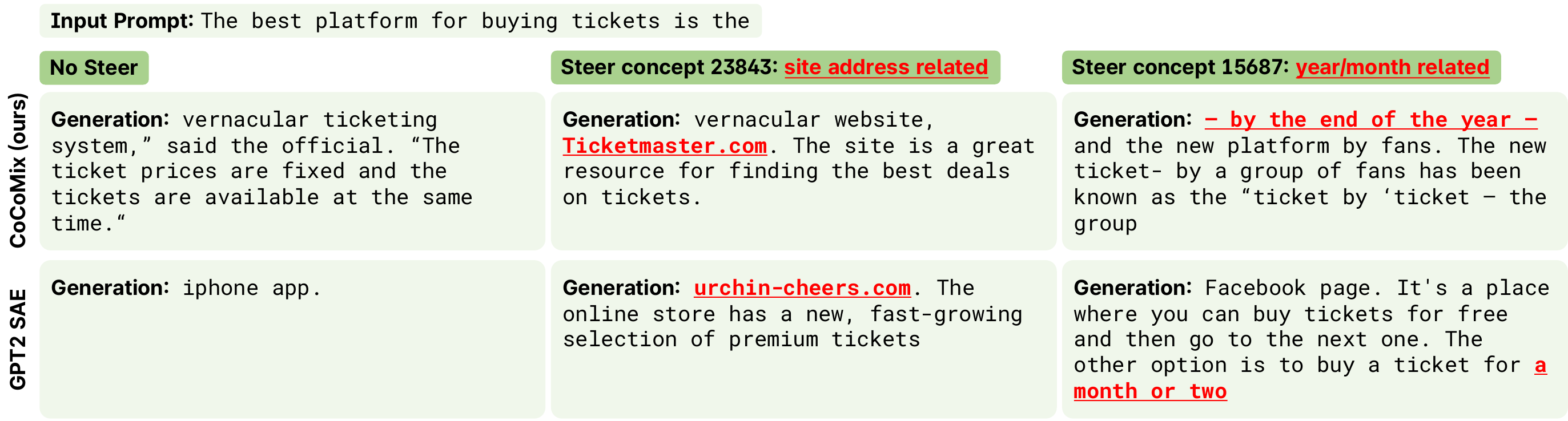}
    \caption{\textbf{Qualitative demonstration of the concept steering effect.} \sname and GPT2 models are 350M and 124M parameter transformers, respectively, trained on the OpenWebText dataset. For \sname, we manipulate the predicted concept logit $\rvz$, while for GPT2, we adjust the SAE concept space $\rvc$ by increasing the activation of a specific concept index. This illustrates the impact of targeted concept steering on the respective model outputs.}
    \label{fig:qualitative_analysis}
\end{figure*}

We provide an empirical evaluation of \sname by investigating the following questions:
\begin{itemize}[leftmargin=*,topsep=0.0pt,itemsep=.5pt]
\item Can \sname improve the performance of next token prediction in LLM pretraining? (\autoref{fig:main_curve} and \autoref{fig:downstream})
\item Does \sname show improvement in weak-to-strong supervision setup compared to other knowledge distillation methods? (\autoref{tab:baseline} and \autoref{fig:baseline})
\item Does \sname introduce model interpretability and steerability? (\autoref{fig:qualitative_analysis})
\item How does each proposed component of \sname contribute to the performance? (\autoref{fig:analysis})
\end{itemize}

Before answering each question, we outline the experimental protocol (more details in Appendix \ref{appn:exp}). 

\textbf{Training setup.} We use a pretrained open-source SAE that is trained on the 124M-sized GPT-2  \citep{gao2024scaling}. We consider training \sname with three different numbers of active parameters, including, 68M, 386M, and 1.38B, with a context length of 1024. For the analysis and ablation study, we mainly conducted experiments on the 68M model. Note that this activated parameter count includes those in the new layer (i.e., the concept predictor), whereas for other baselines, we match the same number of active parameters by increasing the hidden state dimension size $d$. \sname utilizes fewer FLOPs than Pause token (as discussed in Section \ref{sec:analysis}) but more FLOPs than NTP, due to the interleaving of continuous concepts. We use the OpenWebText dataset \citep{radford2019language} as the pretraining corpus to use the same distribution used to train $\mathcal{M}_{\mathtt{con}}$. All experiments are conducted with 20B training tokens, except for the main experiment in Section \ref{sec:exp_main}, which uses 200B tokens.

\textbf{Baselines.} We consider the standard pretraining next token prediction (NTP) procedure, and the commonly used distillation method, knowledge distillation (KD; \citealp{hinton2015distilling}), which is widely used in pretraining \citep{gu2024miniplm, sanh2019distilbert,team2024gemma}. We excluded KD baselines that require multiple models (i.e., more than a single teacher model) to be trained \citep{gu2024miniplm}. For KD, we minimize the KL divergence between teacher and student output while balancing the KL term with the NTP loss. Note that generating synthetic datasets for distillation \citep{kim2016sequence,yu2024distilling} is excluded due to inefficiency, as it requires generating more than 20B tokens. 

\textbf{Evaluation setup.} For evaluation, we consider the validation perplexity of the pretraining dataset and 7 downstream tasks to benchmark commonsense reasoning and reading comprehension. This includes LAMBADA \citep{paperno2016lambada}, WikiText-103 \citep{merity2017pointer}, HellaSwag \citep{zellers2019hellaswag}, PIQA \citep{bisk2020piqa}, Social IQA (SIQA; \citealp{sap2019social}), ARC-easy \citep{clark2018think}, and WinoGrande \citep{sakaguchi2020winogrande} datasets. We also consider OpenWebMath \citep{paster2023openwebmath} as a pretraining dataset to demonstrate that concepts learned from a pretrained LLM can still be applied to \sname, even when the model was trained on a different corpus.

\begin{table*}[t!]

\centering\small
\resizebox{\textwidth}{!}{
\begin{NiceTabular}{lccccccccccc}
\CodeBefore
\rectanglecolor{metabg}{5-1}{5-12}
\rectanglecolor{metabg}{8-1}{8-12}
\rectanglecolor{metabg}{11-1}{11-12}
\Body
\toprule
\multirow{2}{*}{Method} & \multirow{2}{*}{\makecell{Total\\ Params}} & OWT & LAMB & Wiki & HellaS & PIQA  & SIQA & Arc-E  & WinoG & Avg & Avg \\ 
& & PPL ($\downarrow$) & PPL ($\downarrow$) & PPL ($\downarrow$) & Acc-n ($\uparrow$) & Acc ($\uparrow$) & Acc ($\uparrow$) & Acc ($\uparrow$) & Acc ($\uparrow$) & PPL ($\downarrow$) & Acc ($\uparrow$) \\
\midrule
NTP & 69M & 25.3 & 107.6 & 52.3 & 27.4 & 59.4 & 36.6 & 39.7 & 50.7 & 61.8 & 42.7 \\
KD & 69M & 25.2 & \textcolor{white}{0}99.3 & 51.0 & 27.4 & $\bm{59.8}$ & 36.2 & $\bm{39.8}$ & 50.7 & 58.5 & 42.8 \\
\sname & 69M & $\bm{24.7}$ & \textcolor{white}{0}$\bm{99.1}$ & $\bm{50.9}$ & $\bm{27.6}$ & {59.5} & $\bm{37.2}$ & {39.3} & $\bm{51.0}$ & $\bm{58.2}$ & $\bm{42.9}$ \\
\midrule
NTP & 386M & 16.3 & 26.3 & 29.9 & 33.6 & $\bm{64.1}$ & 38.4 & 47.3 & 50.9 & 24.2 & 46.8 \\
KD & 386M & 16.4 & 24.6 & $\bm{29.1}$ & 33.6 & 63.6 & 38.0 & $\bm{48.5}$ & 51.4 & 23.4 & 47.0 \\ 
\sname & 386M & $\bm{15.9}$ & $\bm{19.3}$ & $\bm{29.1}$ & $\bm{34.7}$ & 63.6 & $\bm{39.2}$ & {47.9} & $\bm{52.3}$ & $\bm{21.4}$ & $\bm{47.5}$ \\
\midrule
NTP & 1.38B & 14.3 & 16.6 & 25.0 & 38.1 & 66.1 & 38.9	& 50.5 & 50.0 & 18.6 & 48.7 \\
KD & 1.38B & 14.2 & 16.6 & $\bm{24.9}$	& 37.4 & 66.7 & 39.0 & 50.1 & 52.3 & 18.5 & 49.1 \\
\sname & 1.38B & $\bm{13.9}$ & $\bm{15.4}$ & $\bm{24.9}$ & $\bm{38.4}$ & $\bm{66.9}$ & $\bm{39.5}$ & $\bm{50.8}$ & $\bm{53.0}$ & $\bm{18.1}$ & $\bm{49.7}$ \\
\bottomrule
\end{NiceTabular}
}
\caption{
\textbf{\sname vs. Next Token Prediction (NTP) vs. Knowledge Distillation (KD).} We report performance on the OpenWebText (OWT) training set, as well as downstream tasks, including LAMBADA (LAMB), WikiText-103 (Wiki), HellaSwag (HellaS), PIQA, Social Interaction QA (SIQA), Arc-Easy (Arc-E), and WinoGrande (WinoG). We train three different sizes of models where 124M model is used as a teacher. All models are trained on 20B tokens sampled from the OpenWebText dataset. The bold indicates the best result.
}
\label{tab:baseline}
\end{table*}

\subsection{Main Results}
\label{sec:exp_main}

In this section, we illustrate two core results: i) the comparison with NTP on a relatively large-scale pretraining setup and ii) the comparison with KD baseline, especially on weak-to-strong supervision scenarios where concepts extracted from a small model are used to guide a larger model. 

\textbf{Improving NTP with \sname at scale.}
We first present the main result by applying \sname to the NTP. Here, we consider training NTP and \sname on 200B tokens. As shown in \autoref{fig:downstream}, \sname consistently and significantly improves the performance of overall downstream tasks on various model sizes. Our results also indicate that larger models (e.g., 386M and 1.38B) can benefit from using concepts extracted from a smaller 124M model, showing effective weak-to-strong supervision. Moreover, as shown in \autoref{fig:main_curve}, \sname consistently improves the performance over the NTP on a billion-scale model. For instance, \sname achieves similar performance to NTP while using 21.5\% fewer tokens, demonstrating high sample efficiency. Finally, it is worth noting that the performance gain of using \sname is increasing over the training steps, demonstrating strong generalization performance.

\textbf{Comparison with KD baseline.} We also compare \sname with KD baseline across multiple scenarios, including (1) a stronger teacher model teaching a smaller student model; (2) weak-to-strong supervision, where a weaker teacher teaches a larger student model; and (3) distribution shift, where the student is trained on a corpus different from the teacher’s pretraining distribution. As shown in \autoref{tab:baseline}, \sname demonstrates improvements over KD in all considered model configurations. In particular, \sname shows significant performance gain in the weak-to-strong supervision setup, e.g., improving average perplexity of 2.8 in 386M, while KD does not show great improvement. This arises because a weaker teacher can introduce noisy or suboptimal knowledge, especially when the student surpasses the teacher in capability \citep{rawat2024little}. This trend is also observable in \autoref{fig:baseline}, where models trained with KD fall behind standard training midway through training as the student outpaces the teacher (especially in the distribution shift scenario). In contrast, \sname selectively utilizes useful concepts, resulting in a consistent performance gain.

\subsection{Interpretability and Steerability of \sname}
Another core advantage of \sname is its interpretability and model steering. Specifically, as the model is trained to predict concepts in its hidden state, we can analyze which concepts it focuses on based on the concept predictions. Furthermore, by amplifying the magnitude of the predicted concept $\rvz_t$, one can control the output generation of the model. Following \citet{templeton2024scaling}, we multiply $\rvz_t$ of a desired concept element by constants ranging from -10 to 10. To verify whether this steerability works as intended, we steer the activation of the same concept in the pretrained model's SAE latent space $\rvc$ and confirm whether the output exhibits the corresponding concept. Here, we use a 386M parameter model trained with \sname, where the pretrained model is GPT-2. As shown in \autoref{fig:qualitative_analysis}, when the concept related to ``website address'' is amplified, both models start generating actual website addresses. This demonstrates that our model has successfully learned the GPT-2 aligned concepts. More examples of steering can be found in Appendix \ref{sec_appn:more_steerability}.

\begin{figure*}[t]
\centering\small
\begin{subfigure}{0.3275\textwidth}
    \includegraphics[width=\textwidth,height=0.75\textwidth]{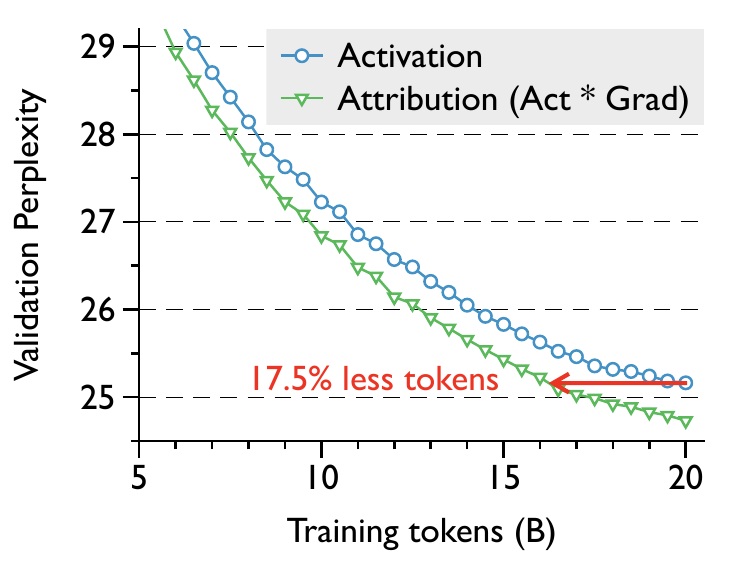}
    \caption{Effectiveness of the attribution score}
    \label{fig:attribution_effect}
\end{subfigure}
\begin{subfigure}{0.3275\textwidth}
    \includegraphics[width=\textwidth,height=0.75\textwidth]{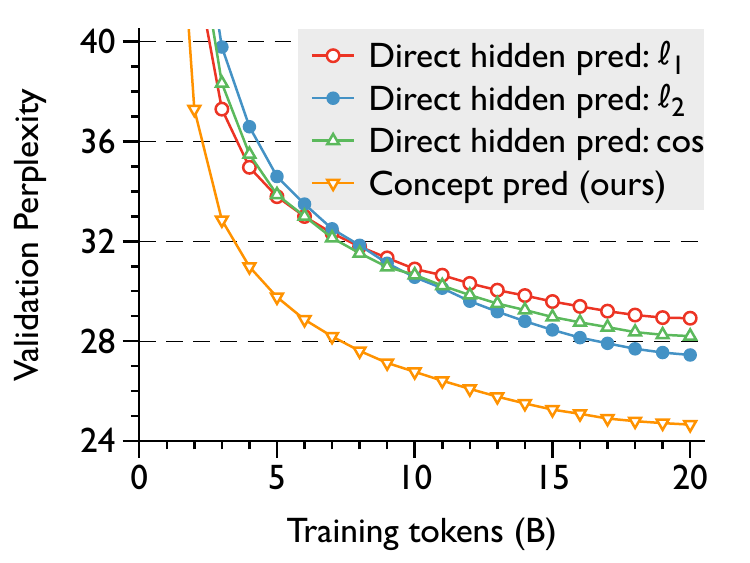}
    \caption{Concept vs. direct hidden state}
    \label{fig:sae_vs_direct_hidden}
\end{subfigure}
\begin{subfigure}{0.3275\textwidth}
    \includegraphics[width=\textwidth,height=0.75\textwidth]{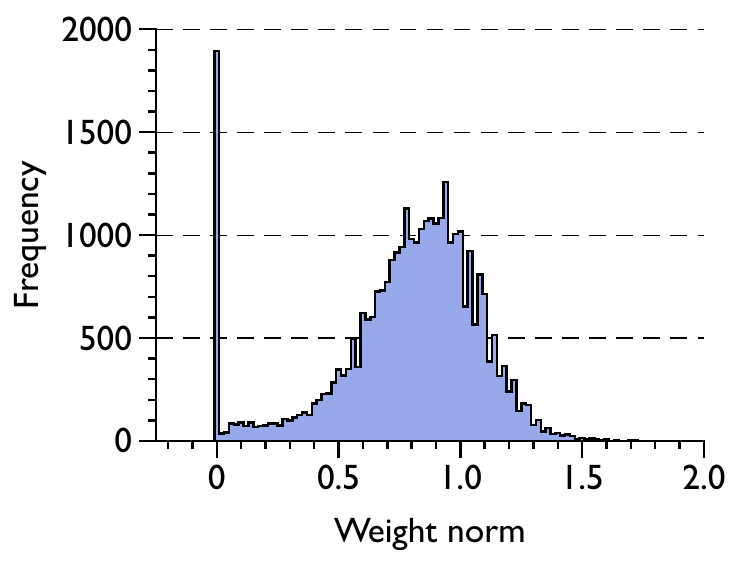}
    \caption{Compression layer's weight analysis}
    \label{fig:compress_weight}
\end{subfigure}
\begin{subfigure}{0.3275\textwidth}
    \includegraphics[width=\textwidth,height=0.75\textwidth]{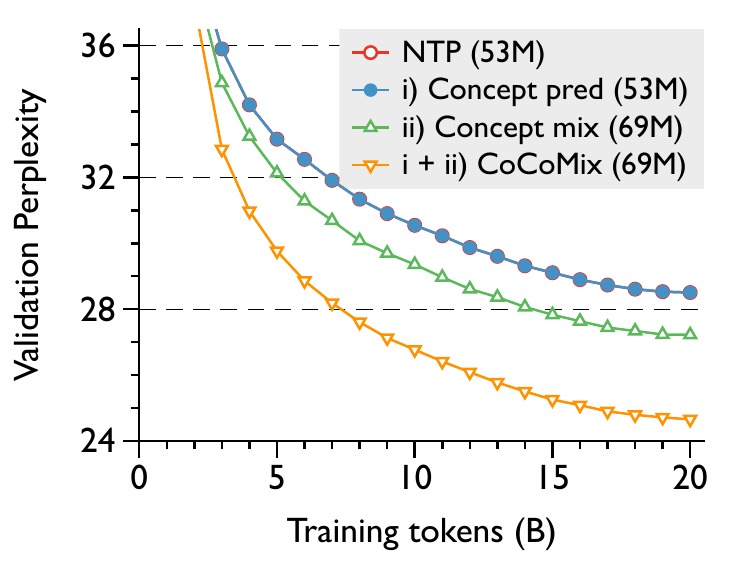}
    \caption{Component analysis}
    \label{fig:component}
\end{subfigure}
\begin{subfigure}{0.3275\textwidth}
    \includegraphics[width=\textwidth,height=0.75\textwidth]{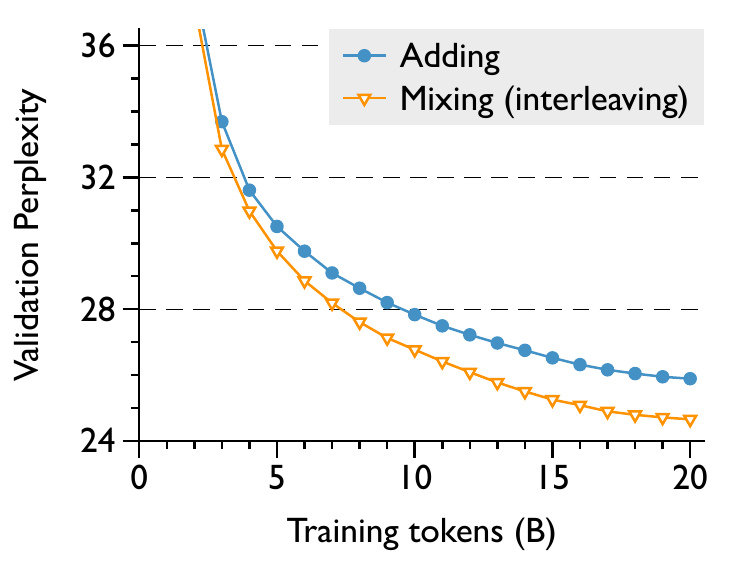}
    \caption{Design choice for concept condition}
    \label{fig:concept_condition}
\end{subfigure}
\begin{subfigure}{0.3275\textwidth}
    \includegraphics[width=\textwidth,height=0.75\textwidth]{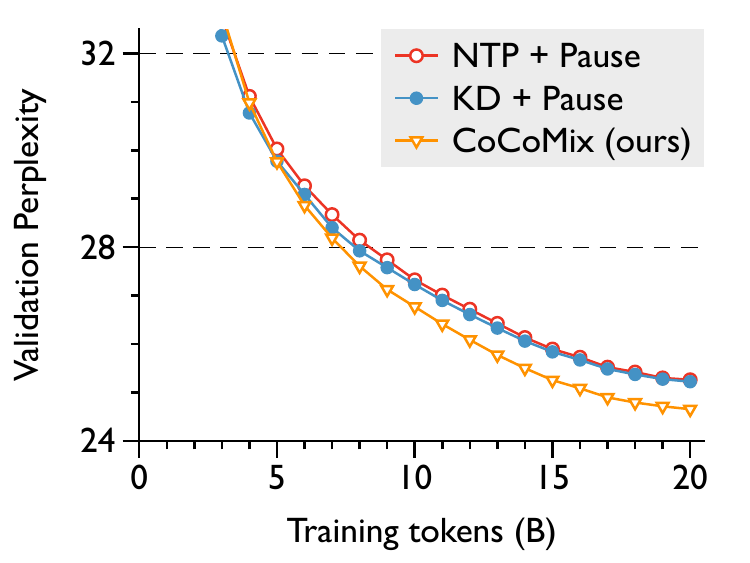}
    \caption{Comparison with Pause Token}
    \label{fig:pause_token}
\end{subfigure}
\caption{
Analysis of \sname: (a) Effectiveness of the attribution score for selecting concepts. (b) Comparison between concept prediction and direct hidden state prediction (i.e., predicting the hidden state with continuous loss rather than discretizing the hidden state with SAE). (c) The sparsity in the compression weight. (d) Component analysis by analyzing the contribution of concept prediction and mixing. (e) Design choices for concept conditioning by comparing adding the concept vector to the original hidden state and mixing (interleaving the concept vector with token hidden representation). (f) Comparison between \sname and the Pause token (i.e., adding learnable tokens). We use a 69M transformer and train on 20B tokens from the OpenWebText dataset.
}
\label{fig:analysis}
\end{figure*}

\subsection{Analysis of  \sname's effectiveness }
\label{sec:analysis}

In this section, we provide a detailed analysis of \sname to validate the effect of each proposed component. Unless otherwise specified, we use a 69M model and train on 20B tokens sampled from the OpenWebText dataset across all methods throughout this section.

\textbf{Effectiveness of the attribution score.} We first analyze whether the attribution score effectively extracts important concepts. To demonstrate this, we train \sname using the activation value $\rvc_t$ for concept extraction, i.e., $\mathcal{L}_{\mathtt{concept}}(\rvc_t)$ in \autoref{eq:concept_pred}, instead of the attribution score $\rva_t$. Remark that the activation value also well reflects the importance of the concept \citep{bricken2023monosemanticity}. As shown in \autoref{fig:attribution_effect}, using attribution scores significantly improves performance, improving sample efficiency by 17.5\% compared to activation value based selection. We believe it will be an interesting future direction to explore other selection criteria for improving \sname's performance or removing undesirable concepts to reduce bias, e.g., selectively removing unsafe concepts for safe LLM pretraining.

\textbf{Comparison with direct hidden state predictions.} To evaluate the importance of predicting the concept extracted from SAE, we compare \sname with direct hidden state prediction strategies (i.e., predict the full hidden state without projecting into the concept space). To have a comparison under the same architecture as \sname, we replace the concept prediction head $M$ with a two-layer multilayer perceptron (MLP), denoted $g(\cdot)$, which predicts the pretrained LLM's hidden state $\rvh^{\mathtt{con}}$ directly from the hidden state of the model $\rvh$. The predicted representation, $g(\rvh)$, is then compressed into a continuous embedding for insertion to have the same architecture as \sname. Here, we use continuous loss including, $\ell_1$, $\ell_2$, and the cosine distance (e.g., $\lvert \rvh^{\mathtt{con}} - g(\rvh) \rvert_2^2$ for $\ell_2$) to predict the hidden state. As shown in \autoref{fig:sae_vs_direct_hidden}, direct hidden state prediction leads to a performance drop. We conjecture this to be due to SAE's ability to decompose the latent into semantically meaningful concepts while predicting all hidden states may include noisy components, emphasizing the effectiveness of our method.

\textbf{Compression layer weight analysis.} Now, we analyze the weight of the compression layer $\rmW$ in \autoref{eq:compress} to show how \sname utilize the predicted concept. To this end, we visualize the $\ell_2$ norm of each concept's weights of 386M \sname: for the weight matrix $\rmW \in \mathbb{R}^{d \times C}$, where $d$ is the hidden dimension and $C$ is the number of concepts, the norm is defined as: $\lVert\rmW_{:,c}\rVert_2 = \sqrt{\sum_{d=1}^D W_{d,c}^2}$. As shown in \autoref{fig:compress_weight}, we found that a portion of concept weights are near zero, e.g., 5.8\% has a norm less than $1\text{e-}2$. This indicates that \sname learns to ignore these concepts when compressing them into a continuous concept if it is not useful. We conjecture such a process enabled \sname for strong weak-to-strong supervision as it learned to ignore ineffective concepts extracted from a weak model.

\textbf{Component Analysis.} We analyze the contributions of each component of our method: (a) concept prediction \autoref{eq:concept_pred}, and (b) concept insertion \autoref{eq:final}. The results in \autoref{fig:component} demonstrate that both components are critical for performance improvement. Specifically, applying the concept prediction loss alone yields a modest reduction in perplexity. However, incorporating concept insertion alongside prediction enhances the effectiveness of the loss, resulting in further performance gains. This highlights the role of insertion in enabling the model to leverage the pretrained LLM's latent reasoning effectively. Notably, while concept insertion increases parameter count, it has a limited impact on performance when used alone, emphasizing the critical role of concept prediction.

\textbf{Continuous concept mixing method.}
We explored two methods for introducing the continuous concept $\hat{\rvc}_t$ into the model’s hidden state $\rvh_t$, referred to as concept conditioning. The first method, {insertion}, insert the continuous concept in front of the token embedding, thus enabling the model to have a mixed input of the token and concept. The other option is {intervention}, which directly modifies the hidden state by adding the concept vector, i.e., $\rvh_t \leftarrow \rvh_t + \hat{\rvc}_t$. As illustrated in \autoref{fig:concept_condition}, both methods enhance pretraining performance (compared to NTP in \autoref{fig:component}), highlighting the importance of concept conditioning, where the insertion method performed better. By introducing a distinct concept vector, the insertion method enables the model to explicitly recognize and effectively utilize concepts during generation, enhancing its overall performance.

\textbf{\sname vs. pause tokens.} Furthermore, we consider an additional baseline that jointly uses pause token \citep{goyal2024think}. Specifically, the pause token uses an additional learnable token that is inserted into the token embedding, enabling the LLM to think more before predicting the next token, which is similar to our continuous concept insertion. To this end, we insert the pause token for every input token on the same hidden state layer as \sname to ensure comparable computation. Moreover, we also consider training the pause token with KD. As shown in \autoref{fig:pause_token}, \sname consistently outperforms the pause token, indicating our inserted (or interleaved) continuous concept indeed consists of useful information to improve the performance.

\section{Related Work}
\label{sec:related}

\textbf{Beyond token-level guidance for language modeling.} While next token prediction remains the standard paradigm for language modeling, recent approaches have begun to explore methods that provide guidance beyond language tokens. For instance, some methods explore a better target, such as leveraging multi-token predictions to capture long context dependencies \citep{gloeckle2024better,liu2024deepseek} or predicting sequence embedding \citep{lee2024semiparametric}. Additionally, methods explore new types of inputs, e.g., using latents \citep{hao2024training} or self-generated thought as inputs \citep{zelikman2024quiet}, which have shown improving reasoning capabilities. Only recently, concept-level modeling using local encoder-decoder architectures has also been explored to represent a language at a higher abstraction level \citep{the2024large}. Other methods add extra tokens in the input space to increase computation at inference time \cite{nye2021show,wei2022chain,goyal2024think,lanchantin2024learning}. In contrast to other works, we propose a pretraining approach that integrates next token prediction with continuous concepts, connecting high level concepts with fine-grained token guidance.

\textbf{Sparse Autoencoders (SAEs).} SAEs extend the autoencoder by enforcing sparsity constraints in the latent space \citep{lee2006efficient}. The features learned by SAEs are often interpretable and disentangled, making them useful across various domains, including language modeling \citep{bricken2023monosemanticity}. Additionally, SAEs have gained attention in mechanistic interpretability due to their ability to capture coherent semantic concepts \citep{marks2024sparse}. This property has enabled practical advancements in identifying and manipulating semantic concepts and facilitating steering for controlled model outputs \citep{lieberum2024gemma}. Among SAE variants, TopK SAEs \citep{makhzani2014k} enforce explicit sparsity using a TopK activation function, demonstrating effectiveness even for large models \citep{gao2024scaling}. In this work, we leverage SAE and, to the best of our knowledge, are the first to apply it to LLM pretraining, achieving strong performance while enhancing the interpretability and controllability of the trained model.

\textbf{Knowledge distillation (KD).} 
Our method can also be related to KD, i.e., transfers the expertise of a teacher model to a student model to enhance performance \citep{hinton2015distilling,zagoruyko2017paying}, as \sname extracts high-level semantic features from a pretrained model which is used to train a base model. Recently, KD for LLMs has garnered increasing attention, leveraging knowledge from a teacher to improve the generative and encoding capabilities of a student \citep{sanh2019distilbert,ko2024distillm}. Especially, applying KD to LLM pretraining remains challenging due to the massive token scales (billions to trillions), forcing most current methods to resort to naive token-level probability matching  \citep{team2024gemma,gu2024miniplm}. Additionally, while pretrained models contain a vast amount of learned information and are thus beneficial to use, reusing knowledge from smaller teacher models remains challenging \citep{burns2023weak}. In this work, we show \sname can even leverage the concept extracted from small models to train a large model showing weak-to-strong supervision.

\section{Conclusion}
\label{sec:con}

We propose \mname, a new LLM pretraining framework that augments next token prediction with continuous concepts. By leveraging concepts extracted from a pretrained SAE as targets, our model predicts both the next token and the associated concept. Predicted concepts are then compressed into a continuous concept vector, which is then mixed into the hidden state. This approach enhances interpretability and controllability by enabling direct probing of the distilled concepts. Experimental results show that \sname consistently improves performance across benchmarks, particularly in challenging generalization scenarios such as weak-to-strong supervision. Future work could explore learning continuous concepts during pretraining without the need for distillation.

\clearpage
\newpage
\bibliographystyle{assets/plainnat}
\bibliography{paper}

\clearpage
\newpage
\beginappendix

\section{Experimental Details}
\label{appn:exp}

\textbf{Architecture details.} Our method and baseline both utilize the GPT-2-based Transformer architecture and tokenizer \citep{radford2019language}, with a context length of 1024. We consider three model sizes, defined by the number of activated parameters: 69M, 386M, and 1.38B. For \sname, the hidden state dimensions $d$ are 512, 1024, and 2028, with 8, 24, and 24 layers, respectively. The number of attention heads is set to 8, 16, and 16 for the three configurations. Baseline models are configured to match the number of activated parameters by employing the same number of layers and attention heads, with hidden state dimensions of 624, 1072, and 2096, respectively. We leverage an open-source pretrained TopK Sparse Autoencoder (SAE) \citep{gao2024scaling} for concept extraction, where $K_{\mathtt{concept}}$ is set to 32 and the concept activation size is 32,768. Consequently, our models introduce an additional $(C + K_{\mathtt{concept}}) \times d$ activated parameters on top of the base Transformer parameters. The GPT-2 model (a teacher model for KD and a concept extraction model for \sname) has 12 layers, a hidden dimension size of 768, and 124M parameters. The concept extraction layer $L_{\mathtt{con}}$ is configured as the 6th middle layer of GPT-2 for all model sizes. For \sname, the concept prediction is done at the 4th layer for the 69M model and the 6th layer for the larger 386M and 1.38B models.

\textbf{Training details.} 
We mainly followed the training details outlined in \citep{brown2020language}. For the main results presented in Section~\ref{sec:method}, models were trained on 200B tokens over approximately 382K training steps, while other experiments were conducted on 20B tokens. We used OpenWebText as the training dataset to match the same training corpus as GPT-2. The learning rate schedule included a warm-up phase over the first 1/300 of the total training steps, followed by a cosine decay to 10\% of the maximum learning rate at the end of training. The maximum learning rates were set to $6\text{e-}4$, $3\text{e-}4$, and $2\text{e-}4$ for the 69M, 386M, and 1.38B models, respectively. The batch sizes were configured to 0.5M, 0.5M, and 1M tokens for the three model sizes. A weight decay of 0.1 was applied across all configurations, and we utilized the AdamW optimizer \citep{loshchilov2017decoupled} with $\beta_1 = 0.9$ and $\beta_2 = 0.95$. For training \sname, the concept prediction loss \autoref{eq:final} was scaled by a hyperparameter $\lambda = 0.1$, and $K_{\mathtt{attri}}$ was set to 4. For the KD baseline, we employed the vanilla KD loss, where the output probabilities of the teacher model $\mathcal{M}_{\mathtt{con}}$ and the student model $\mathcal{M}$ were matched using the Kullback-Leibler (KL) divergence. Specifically, given an input $\rvx$, the KD loss is defined as $-\log \mathcal{M}(x_{t+1}|\rvx) + \lambda_\text{KD} \cdot \mathrm{KL}(\mathcal{M}_{\mathtt{con}}(\cdot|\rvx) \| \mathcal{M}(\cdot|\rvx))$, where $\lambda_\text{KD}=0.1$ consistently demonstrated strong performance.

\section{Additional Results}

\subsection{Detailed Performance During Training}

\begin{figure*}[t]
\centering\small
\vspace{0.1in}
\begin{subfigure}{0.245\textwidth}
    \includegraphics[width=\textwidth,height=0.75\textwidth]{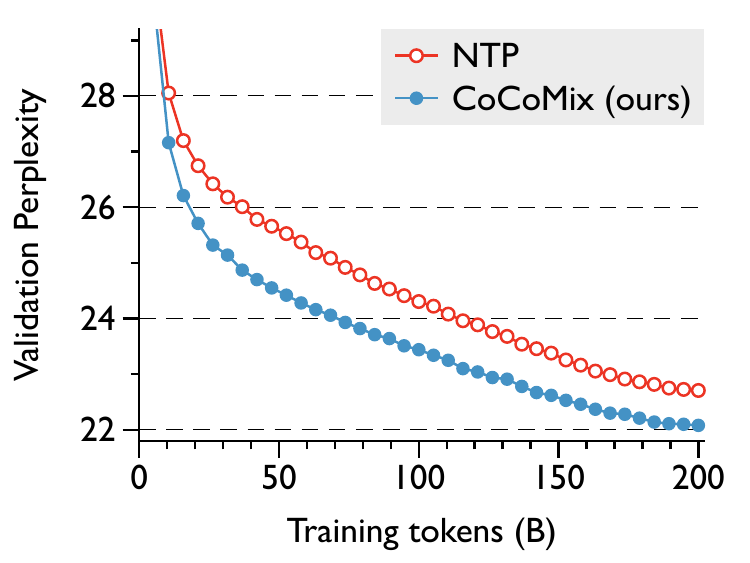}
    \caption{OpenWebText, PPL ($\downarrow$)}
\end{subfigure}
\begin{subfigure}{0.245\textwidth}
    \includegraphics[width=\textwidth,height=0.75\textwidth]{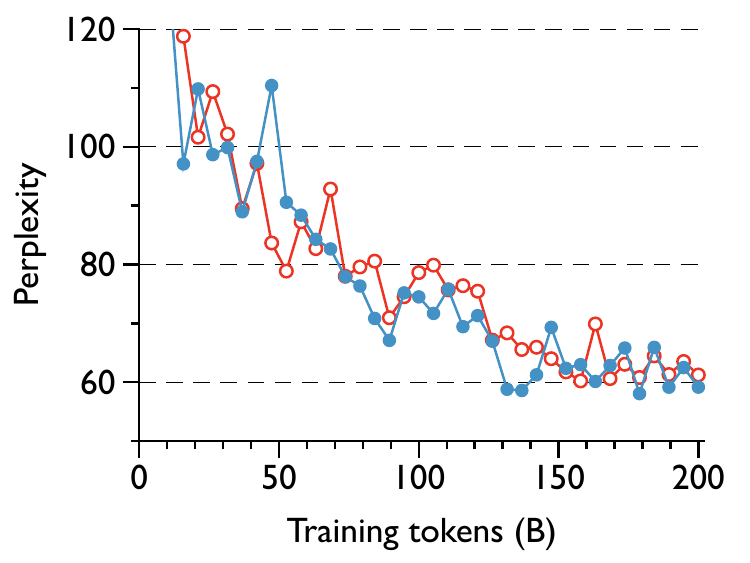}
    \caption{LAMBADA, PPL ($\downarrow$)}
\end{subfigure}
\begin{subfigure}{0.245\textwidth}
    \includegraphics[width=\textwidth,height=0.75\textwidth]{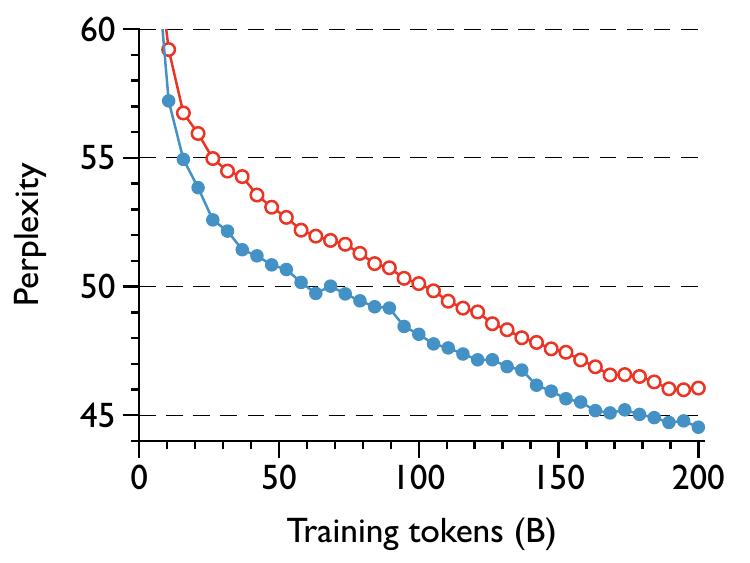}
    \caption{WikiText-103, PPL ($\downarrow$)}
\end{subfigure}
\begin{subfigure}{0.245\textwidth}
    \includegraphics[width=\textwidth,height=0.75\textwidth]{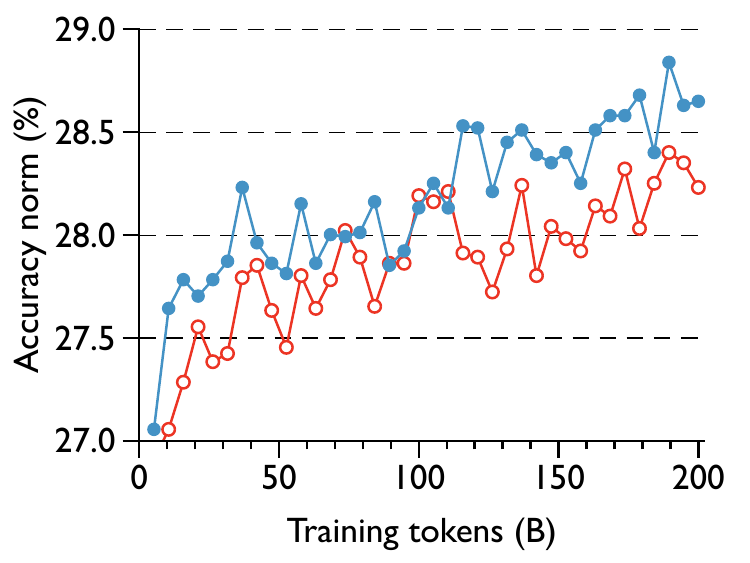}
    \caption{HellaSwag, Acc-n ($\uparrow$)}
\end{subfigure}
\begin{subfigure}{0.245\textwidth}
    \includegraphics[width=\textwidth,height=0.75\textwidth]{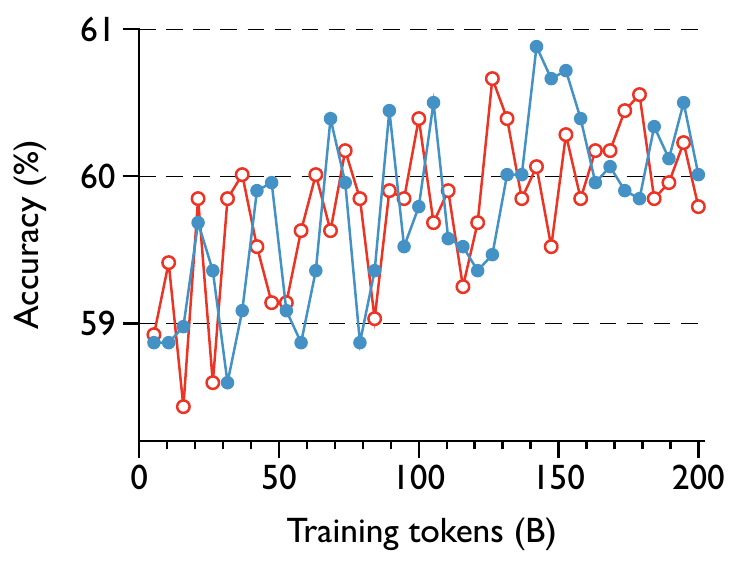}
    \caption{PIQA, Acc ($\uparrow$)}
\end{subfigure}
\begin{subfigure}{0.245\textwidth}
    \includegraphics[width=\textwidth,height=0.75\textwidth]{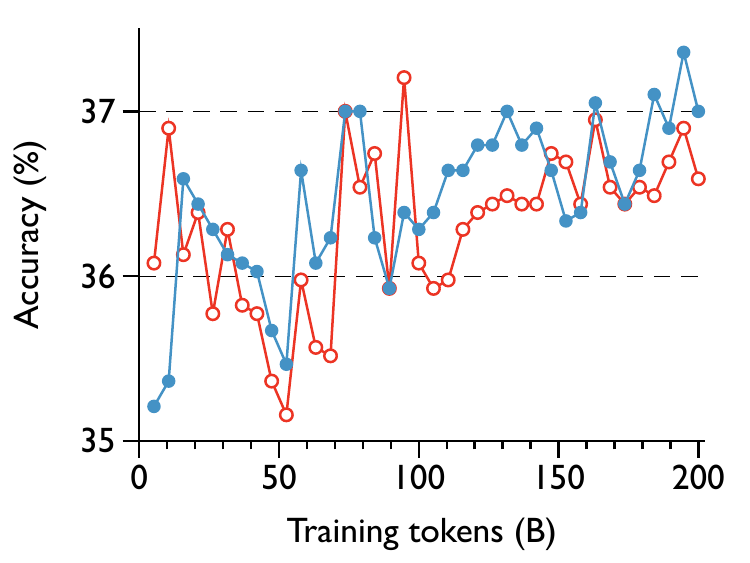}
    \caption{SIQA, Acc ($\uparrow$)}
\end{subfigure}
\begin{subfigure}{0.245\textwidth}
    \includegraphics[width=\textwidth,height=0.75\textwidth]{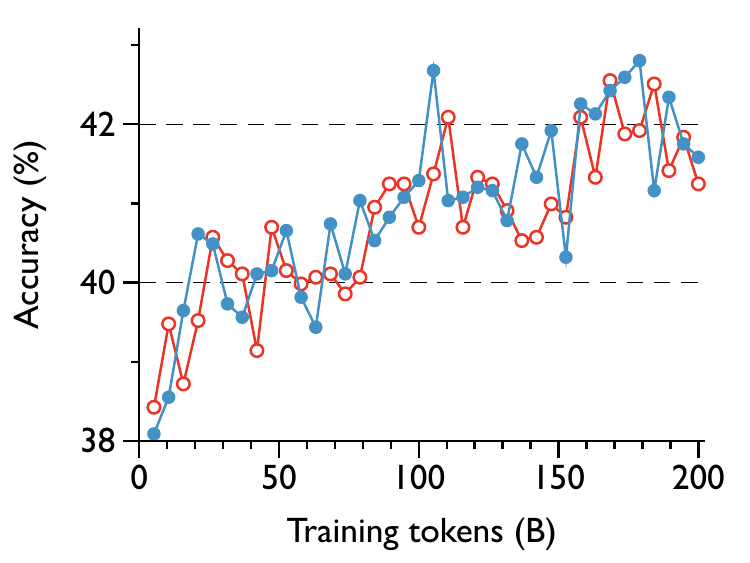}
    \caption{Arc-Easy, Acc ($\uparrow$)}
\end{subfigure}
\begin{subfigure}{0.245\textwidth}
    \includegraphics[width=\textwidth,height=0.75\textwidth]{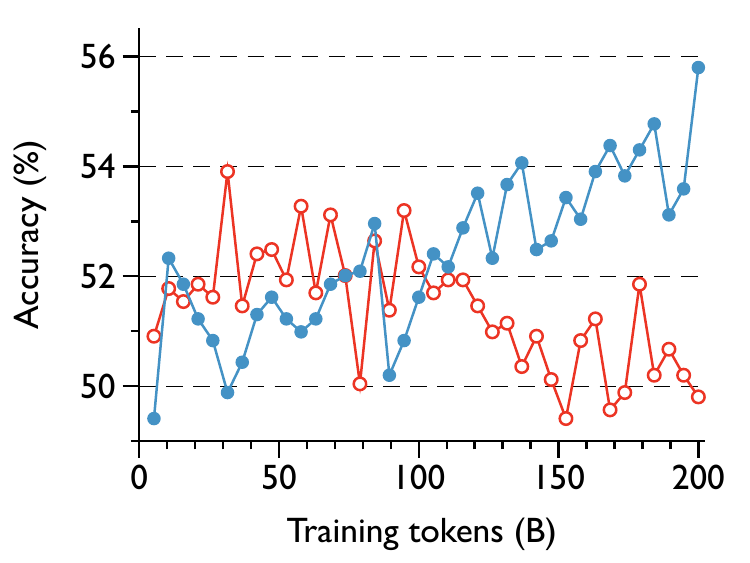}
    \caption{WinoGrande, Acc ($\uparrow$)}
\end{subfigure}
\caption{
\textbf{\sname vs. NTP performance at different training checkpoints on 69M parameter model.} Each model is trained on the 200B tokens sampled from the OpenWebText dataset. The plot shows the result of (a) OpenWebText, (b) LAMBADA, (c) WikiText-103, (d) HellaSwag, (e) PIQA, (f) SIQA, (g) Arc-Easy, and (h) WinoGrande datasets. We use the concepts extracted from a 124M-sized model for training \sname.
}
\label{fig:downstream_curve_69m}
\end{figure*}

\begin{figure*}[t]
\centering\small

\begin{subfigure}{0.245\textwidth}
    \includegraphics[width=\textwidth,height=0.75\textwidth]{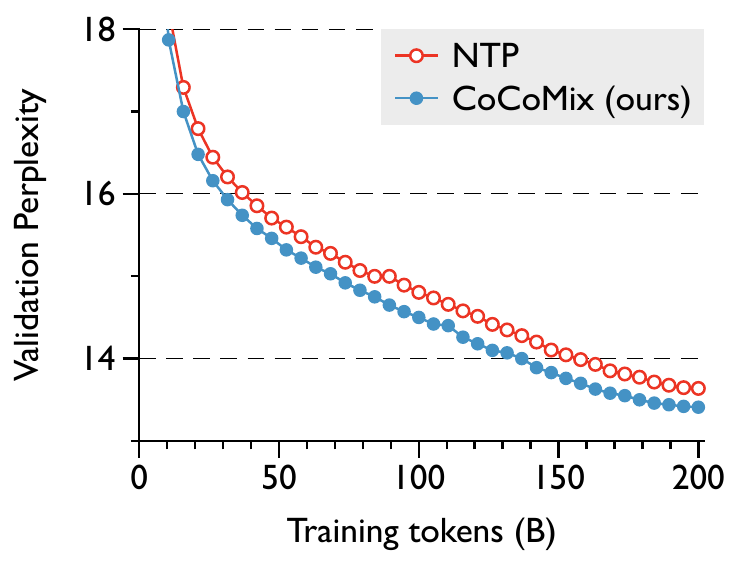}
    \caption{OpenWebText, PPL ($\downarrow$)}
\end{subfigure}
\begin{subfigure}{0.245\textwidth}
    \includegraphics[width=\textwidth,height=0.75\textwidth]{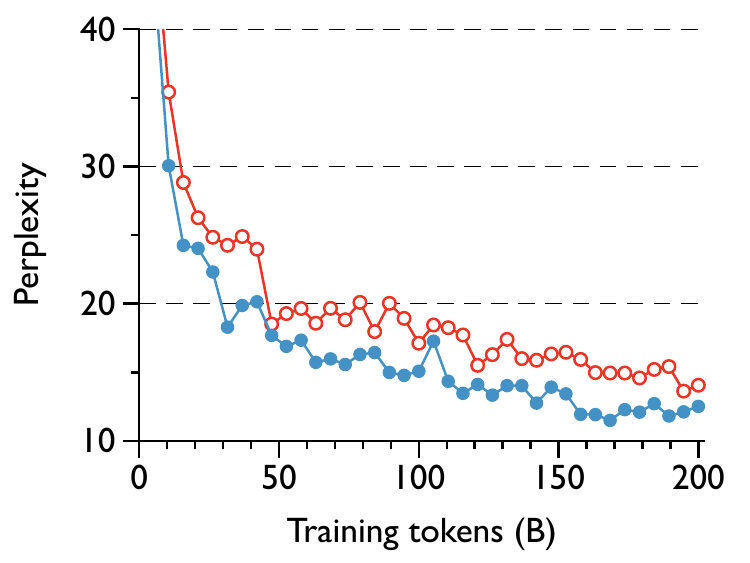}
    \caption{LAMBADA, PPL ($\downarrow$)}
\end{subfigure}
\begin{subfigure}{0.245\textwidth}
    \includegraphics[width=\textwidth,height=0.75\textwidth]{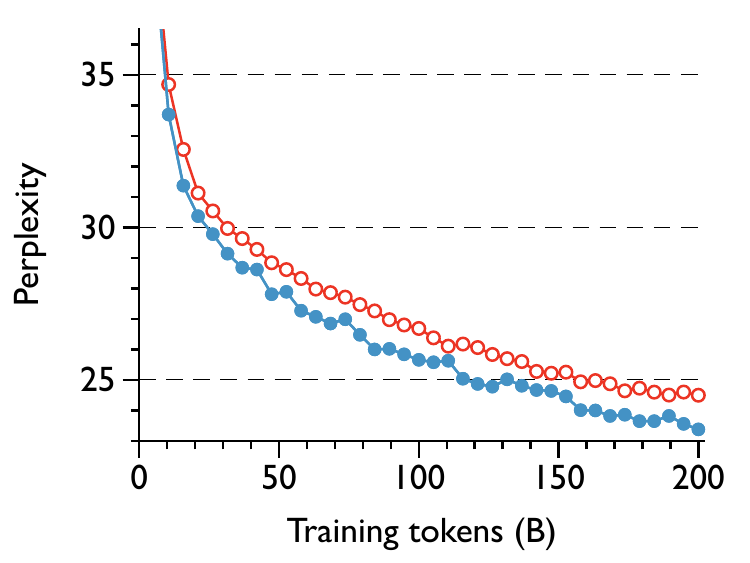}
    \caption{WikiText-103, PPL ($\downarrow$)}
\end{subfigure}
\begin{subfigure}{0.245\textwidth}
    \includegraphics[width=\textwidth,height=0.75\textwidth]{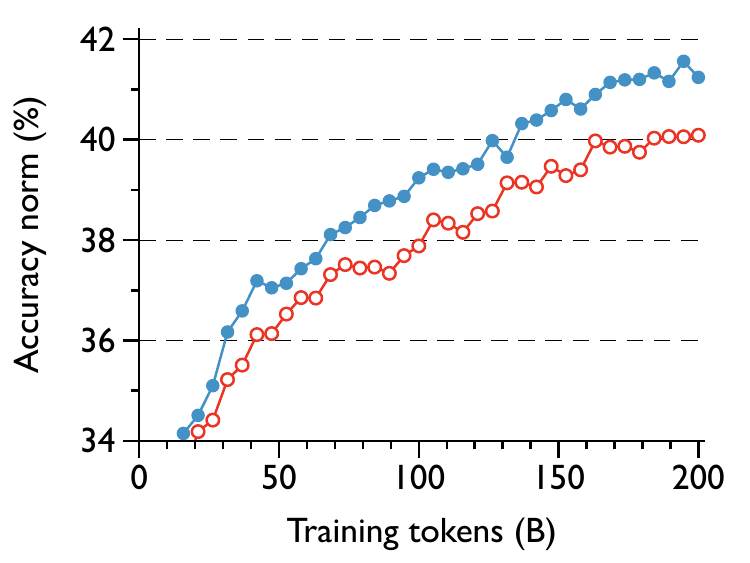}
    \caption{HellaSwag, Acc-n ($\uparrow$)}
\end{subfigure}
\begin{subfigure}{0.245\textwidth}
    \includegraphics[width=\textwidth,height=0.75\textwidth]{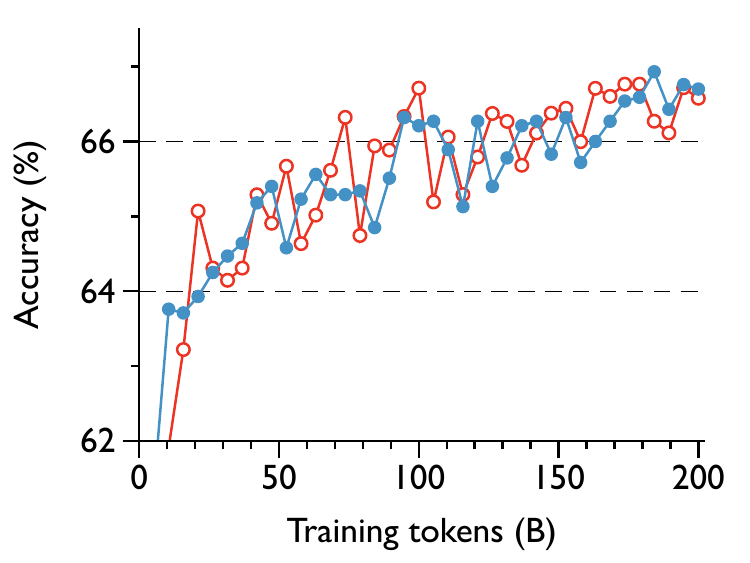}
    \caption{PIQA, Acc ($\uparrow$)}
\end{subfigure}
\begin{subfigure}{0.245\textwidth}
    \includegraphics[width=\textwidth,height=0.75\textwidth]{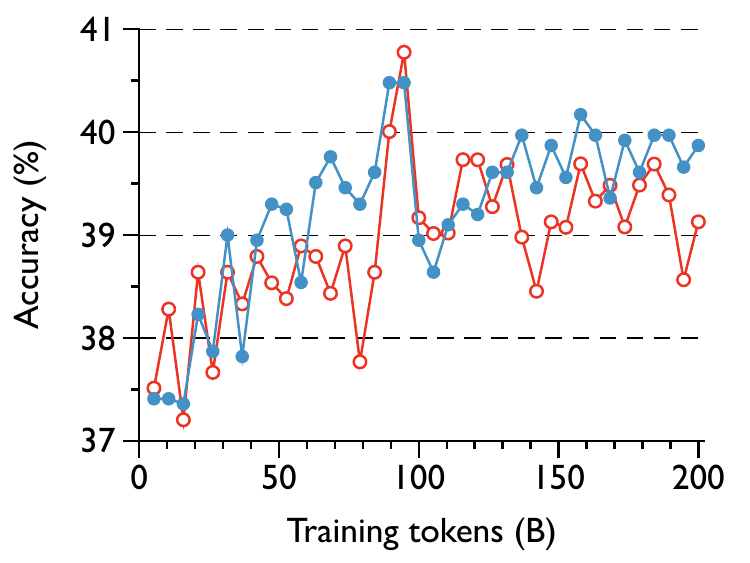}
    \caption{SIQA, Acc ($\uparrow$)}
\end{subfigure}
\begin{subfigure}{0.245\textwidth}
    \includegraphics[width=\textwidth,height=0.75\textwidth]{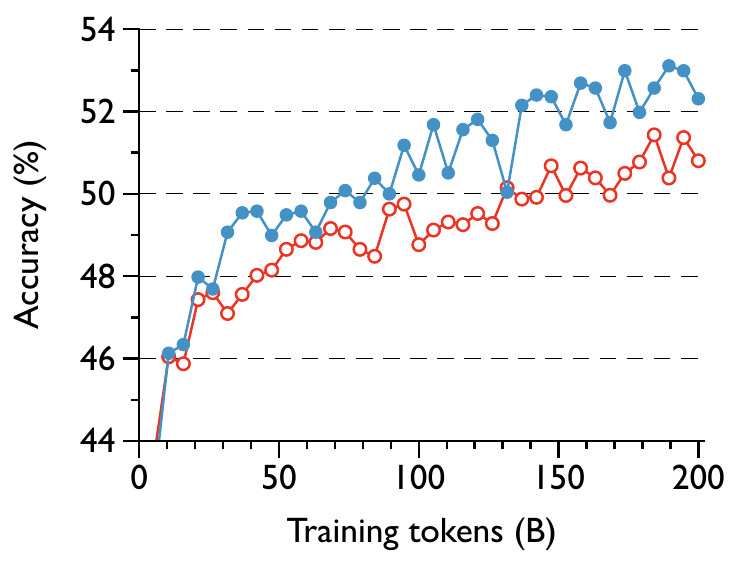}
    \caption{Arc-Easy, Acc ($\uparrow$)}
\end{subfigure}
\begin{subfigure}{0.245\textwidth}
    \includegraphics[width=\textwidth,height=0.75\textwidth]{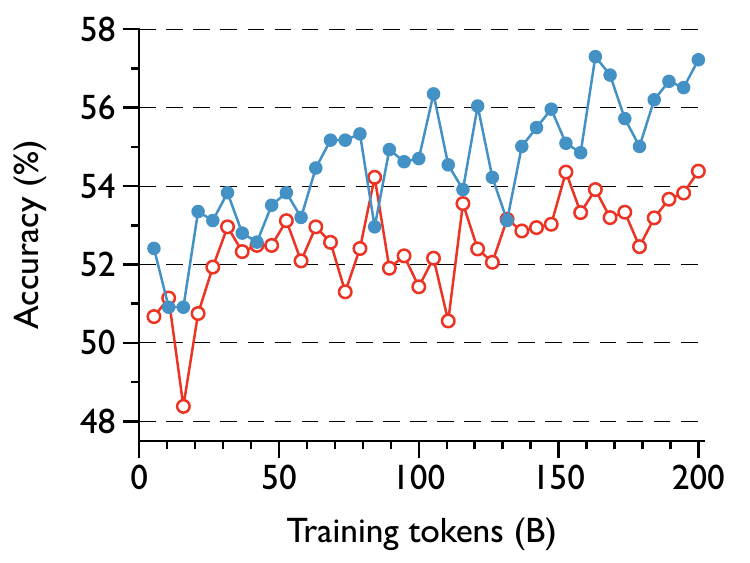}
    \caption{WinoGrande, Acc ($\uparrow$)}
\end{subfigure}
\caption{
\textbf{\sname vs. NTP performance at different training checkpoints on 368M parameter model.} Each model is trained on the 200B tokens sampled from the OpenWebText dataset. The plot shows the result of (a) OpenWebText, (b) LAMBADA, (c) WikiText-103, (d) HellaSwag, (e) PIQA, (f) SIQA, (g) Arc-Easy, and (h) WinoGrande datasets. We use the concepts extracted from a 124M-sized model for training \sname.
}

\label{fig:downstream_curve_386m}
\end{figure*}

\begin{figure*}[t]
\centering\small

\begin{subfigure}{0.245\textwidth}
    \includegraphics[width=\textwidth,height=0.75\textwidth]{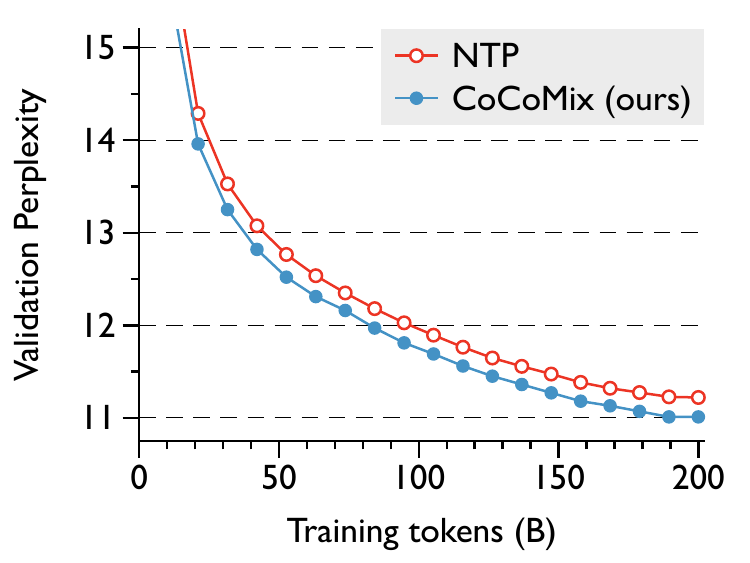}
    \caption{OpenWebText, PPL ($\downarrow$)}
\end{subfigure}
\begin{subfigure}{0.245\textwidth}
    \includegraphics[width=\textwidth,height=0.75\textwidth]{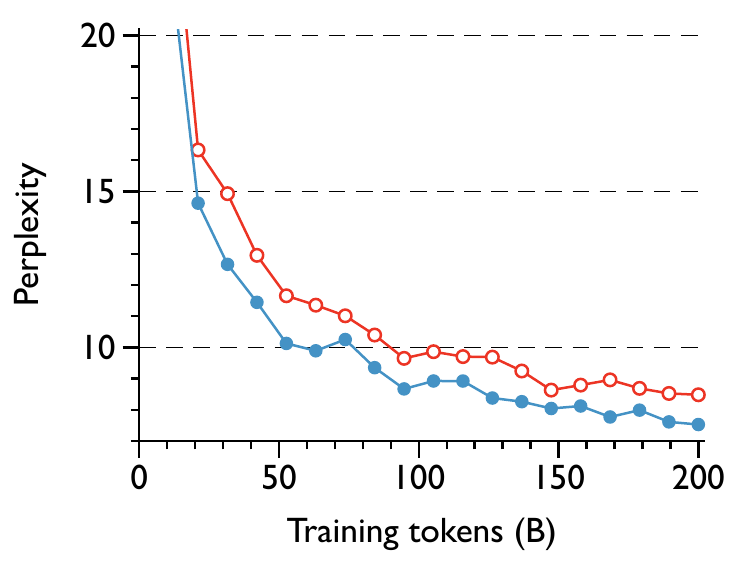}
    \caption{LAMBADA, PPL ($\downarrow$)}
\end{subfigure}
\begin{subfigure}{0.245\textwidth}
    \includegraphics[width=\textwidth,height=0.75\textwidth]{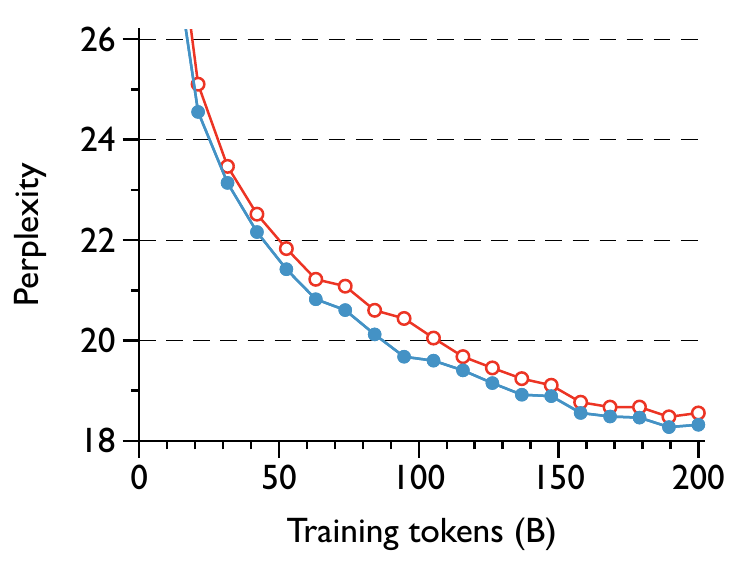}
    \caption{WikiText-103, PPL ($\downarrow$)}
\end{subfigure}
\begin{subfigure}{0.245\textwidth}
    \includegraphics[width=\textwidth,height=0.75\textwidth]{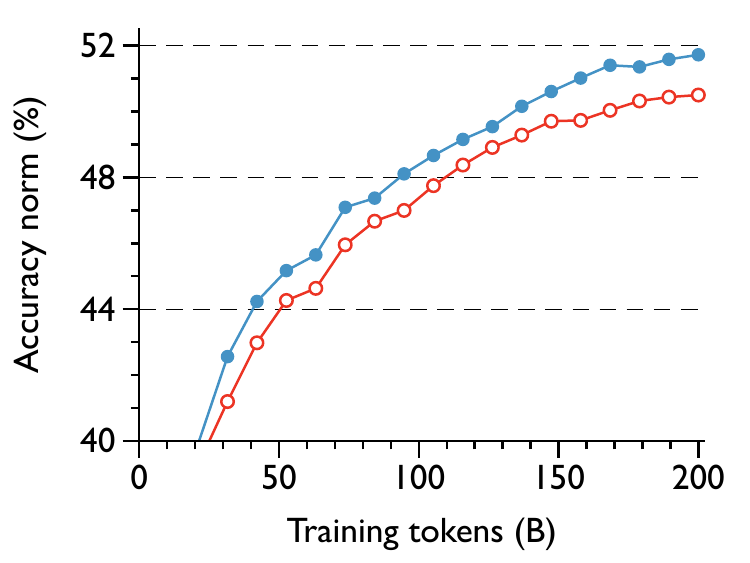}
    \caption{HellaSwag, Acc-n ($\uparrow$)}
\end{subfigure}
\begin{subfigure}{0.245\textwidth}
    \includegraphics[width=\textwidth,height=0.75\textwidth]{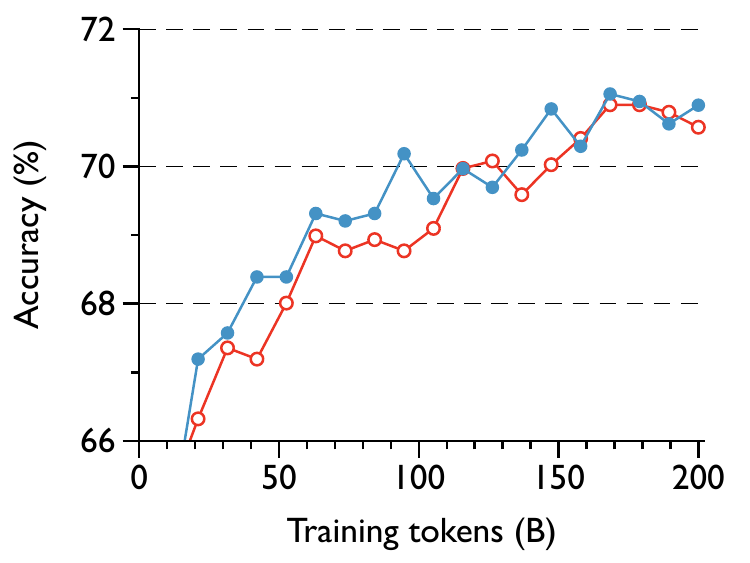}
    \caption{PIQA, Acc ($\uparrow$)}
\end{subfigure}
\begin{subfigure}{0.245\textwidth}
    \includegraphics[width=\textwidth,height=0.75\textwidth]{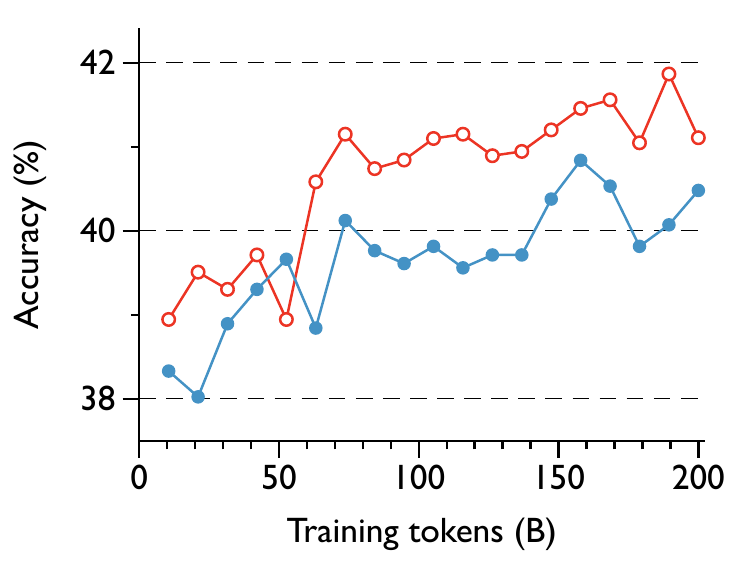}
    \caption{SIQA, Acc ($\uparrow$)}
\end{subfigure}
\begin{subfigure}{0.245\textwidth}
    \includegraphics[width=\textwidth,height=0.75\textwidth]{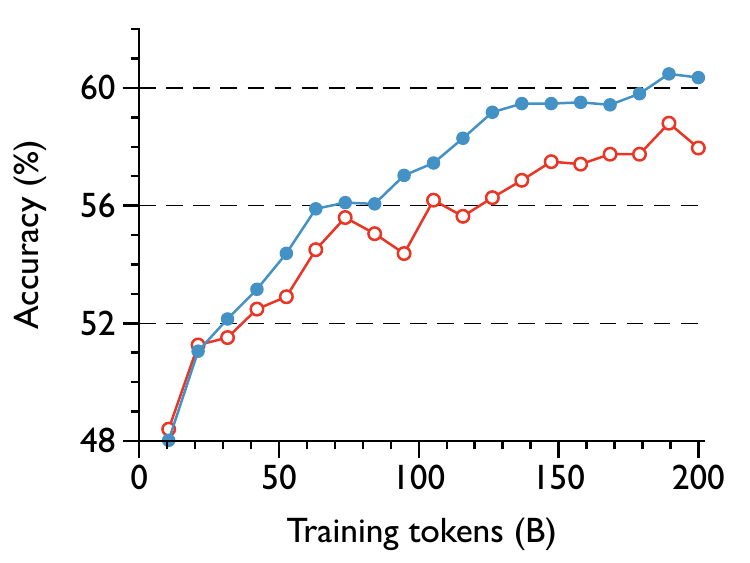}
    \caption{Arc-Easy, Acc ($\uparrow$)}
\end{subfigure}
\begin{subfigure}{0.245\textwidth}
    \includegraphics[width=\textwidth,height=0.75\textwidth]{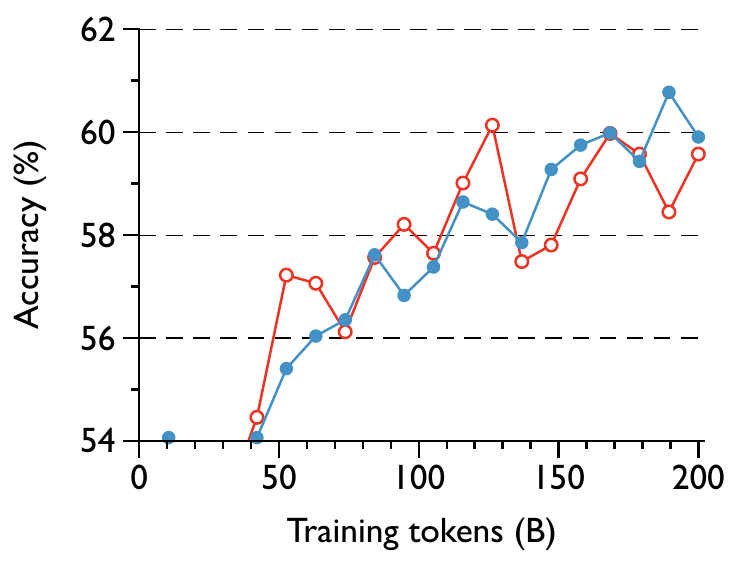}
    \caption{WinoGrande, Acc ($\uparrow$)}
\end{subfigure}
\caption{
\textbf{\sname vs. NTP performance at different training checkpoints on 1.38B parameter model.} Each model is trained on the 200B tokens sampled from the OpenWebText dataset. The plot shows the result of (a) OpenWebText, (b) LAMBADA, (c) WikiText-103, (d) HellaSwag, (e) PIQA, (f) SIQA, (g) Arc-Easy, and (h) WinoGrande datasets. We use the concepts extracted from a 124M-sized model for training \sname.
}
\label{fig:downstream_curve_1b}
\end{figure*}

In this section, we present the performance tracking during training on 200B tokens, including validation perplexity and the perplexity and accuracy of various downstream tasks, including LAMBADA, WikiText-103, HellaSwag, PIQA, SIQA, Arc-Easy, and WinoGrande datasets. As shown in \autoref{fig:downstream_curve_69m}, \autoref{fig:downstream_curve_386m}, and \autoref{fig:downstream_curve_1b}, we compare \sname with the next token prediction (NTP) across different active parameter sizes: 69M, 386M, and 1.38B. In most of the graphs, \sname consistently demonstrates performance gains. Notably, our results show that \sname achieves stable improvements in perplexity across all tasks. Furthermore, \sname exhibits sample efficiency; for instance, in the 1.38B model, \sname reaches the same OpenWebText validation perplexity as NTP while requiring approximately 43B fewer tokens (a 21.5\% improvement in token efficiency).

\subsection{Additional Steerability Results}
\label{sec_appn:more_steerability}

\begin{figure*}[!ht]
    \centering
    \begin{subfigure}{\textwidth}
        \centering
        \includegraphics[width=\linewidth]{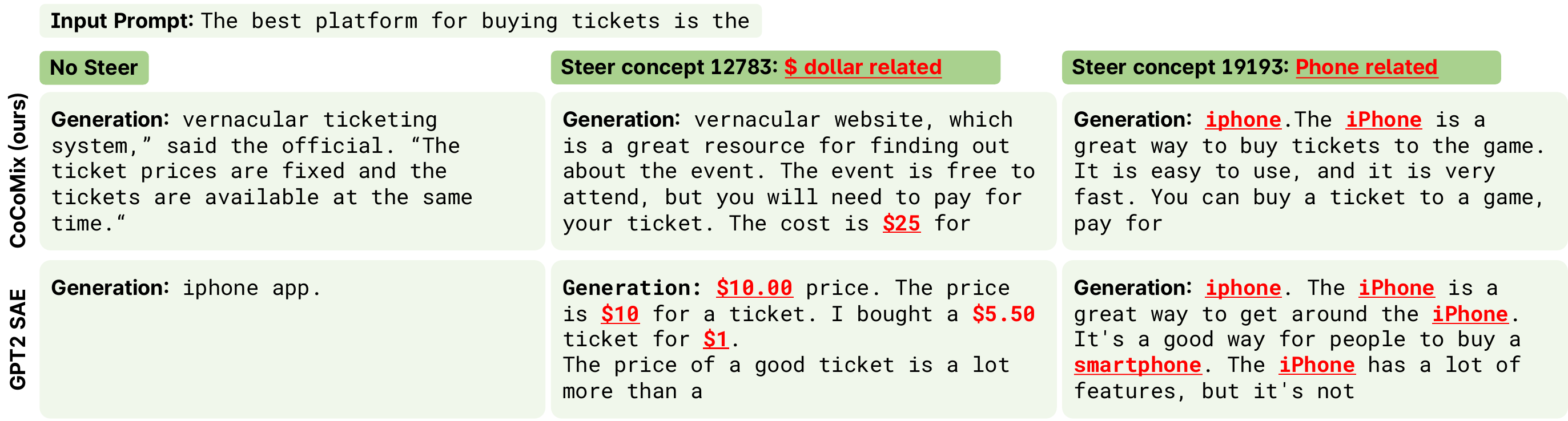}
    \caption{Main figure prompt, `\$ dollar' and `Phone' concept}
    \end{subfigure}
    \begin{subfigure}{\textwidth}
        \centering
        \includegraphics[width=\linewidth]{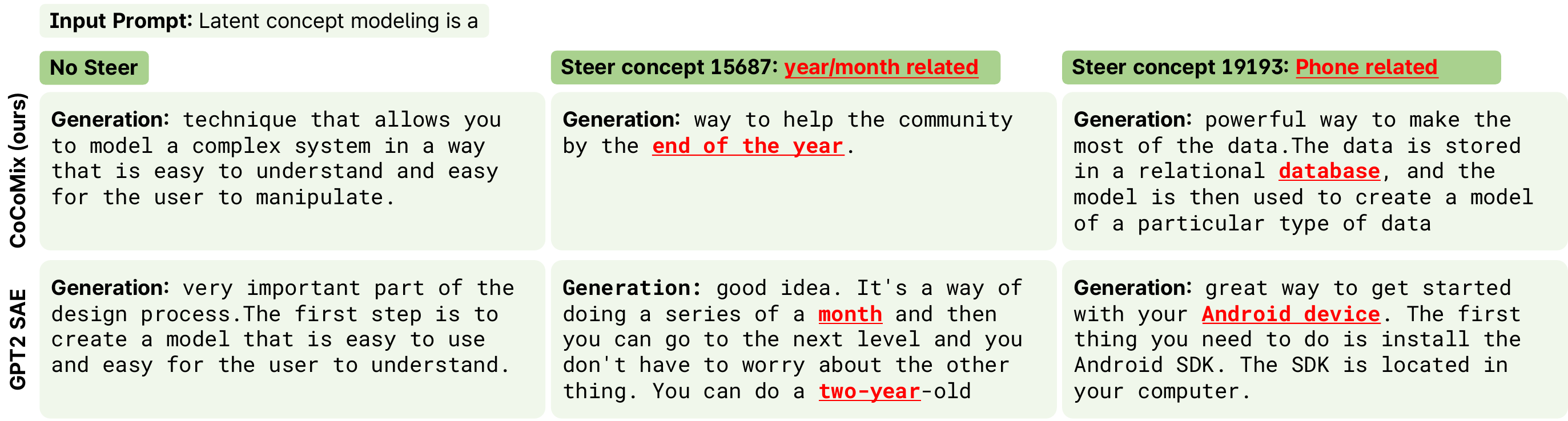}
    \caption{New prompt, `year/month' and `Phone' concept}
    \end{subfigure}
    \begin{subfigure}{\textwidth}
        \centering
        \includegraphics[width=\linewidth]{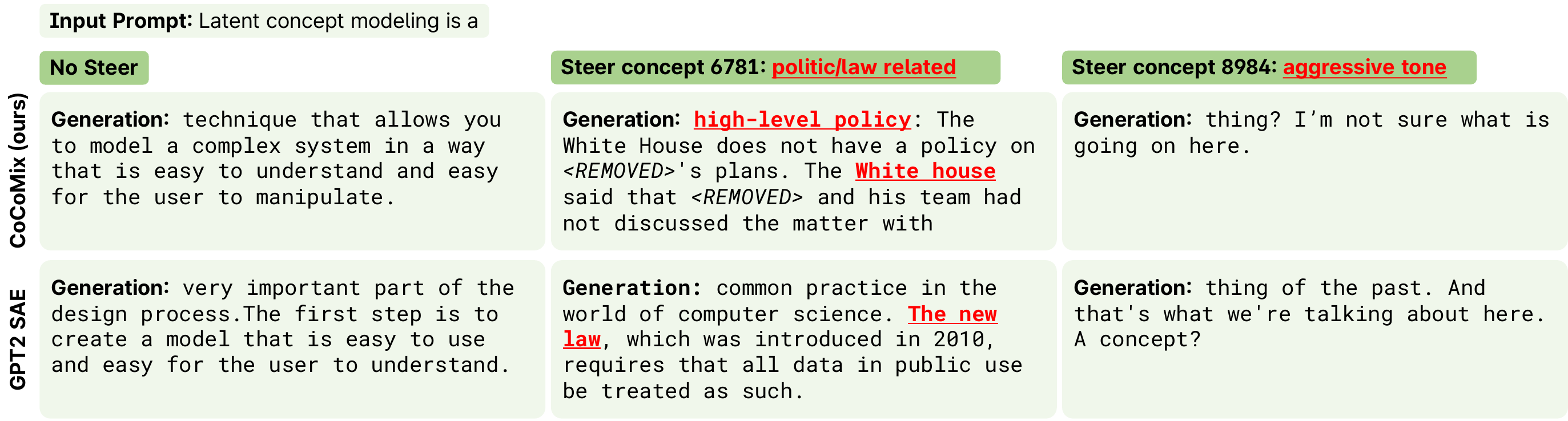}
    \caption{New prompt, `Politic/law' and `aggressive tone' concept}
    \end{subfigure}
    \caption{More qualitative demonstration of the concept steering effect. \sname and GPT2 models are 350M and 124M parameter transformers, respectively. For \sname, we manipulate the predicted logit $\rvz$, while for GPT2, we adjust the SAE concept space $\rvc$ by increasing the activation of a specific concept index}
    \label{fig:appn_qualitative_analysis}
\end{figure*}

To further analyze the steerability enabled by \sname, we conducted experiments using both the same prompt as in the main figure and a new prompt (in \autoref{fig:appn_qualitative_analysis}). For consistency, we first applied steering on additional concepts identified during our analysis—``\$ dollar'' and ``Phone''—using the same prompt as in \autoref{fig:qualitative_analysis}. These experiments confirmed that the model could effectively modulate its output based on these newly identified concepts, producing coherent and concept-aligned generations. Next, to verify whether the identified concepts generalize to different contexts, we experimented with a new prompt: \textit{``Latent concept modeling is a''} and steered the model using the previously identified concepts ``month/year'' and ``Phone.'' The results showed that the model successfully reproduced outputs aligned with these concepts, further supporting the robustness of our method. Additionally, we explored whether new concepts could be identified and steered using the same prompt. In this case, we identified two new concepts: ``politics/law'' and ``aggressive tone.'' Steering the model with these new concepts demonstrated that the outputs could be effectively controlled to exhibit characteristics aligned with the corresponding concepts. These findings further highlight the flexibility and interpretability of our approach.

\end{document}